\documentclass{article}

\usepackage{arxiv}

\usepackage[utf8]{inputenc} 
\usepackage[T1]{fontenc}    
\usepackage{hyperref}       
\usepackage{url}            
\usepackage{booktabs}       
\usepackage{amsfonts}       
\usepackage{nicefrac}       
\usepackage{microtype}      
\usepackage{graphicx}
\usepackage{hyperref}
\usepackage{multirow}
\usepackage{amsmath}
\usepackage{amssymb}
\usepackage{enumitem}   
\usepackage{adjustbox}
\usepackage{subcaption}
\usepackage{xspace}
\usepackage[backend=biber]{biblatex}
\usepackage[nomargin,inline,draft]{fixme}
\fxsetup{theme=color,mode=multiuser}
\FXRegisterAuthor{j}{jla}{\color{blue}JLA}
\FXRegisterAuthor{r}{rdm}{\color{red}RdM}

\newcommand{\no}{NO\textsubscript{2}\xspace}

\title{SOCAIRE: Forecasting and Monitoring Urban Air Quality in Madrid}

\author{
  Rodrigo de Medrano \\
  Artificial Intelligence Department \\
  Universidad Nacional de Educación a Distancia --- UNED \\
  Madrid, Spain \\
  \texttt{rdemedrano@dia.uned.es} \\

  \And
  Víctor de Buen Remiro \\
  Inverence \\
  Madrid, Spain \\
  \texttt{victor.debuen@inverence.com} \\

  \And
  José L. Aznarte \\
  Artificial Intelligence Department \\
  Universidad Nacional de Educación a Distancia --- UNED \\
  Madrid, Spain \\
  \texttt{jlaznarte@dia.uned.es} \\
  
}

\begin{document}
\maketitle

\begin{abstract}
  Air quality has become one of the main issues in public health and urban
  planning management, due to the proven adverse effects of high pollutant
  concentrations. Considering the mitigation measures that cities all over the
  world are taking in order to face frequent low air quality episodes, the
  capability of foreseeing future pollutant concentrations is of great
  importance. Through this paper, we present SOCAIRE, an operational tool based
  on a Bayesian and spatiotemporal ensemble of neural and statistical nested
  models. SOCAIRE integrates endogenous and exogenous information in order to
  predict and monitor future distributions of the concentration for several
  pollutants in the city of Madrid. It focuses on modeling each and every
  available component which might play a role in air quality: past
  concentrations of pollutants, human activity, numerical pollution estimation,
  and numerical weather predictions. This tool is currently in operation in
  Madrid, producing daily air quality predictions for the next 48 hours and
  anticipating the probability of the activation of the measures included in the
  city's official air quality \no protocols through probabilistic inferences
  about compound events.
\end{abstract}


\section{Introduction}
\label{S1}

During the last decades, an increasing number of studies point out that degraded
air quality is a major problem in cities around the world
\cite{martuzzi2006health, heroux_quantifying_2015}. While there is general
consensus that it causes health problems \cite{kim_review_2015,
  ozkaynak2009summary}, how dangerous it can be is still a matter of debate
\cite{sellier_health_2014}. Even in the best case scenario, this seemingly
endemic issue affecting the life in big cities is already considered one of the
main causes of both direct and indirect mortality
\cite{badyda_ambient_2017}.

Of all the tools and systems that help fight pollution, the prediction of future
pollutant concentrations or levels is of principal importance for air quality
management and control \cite{bai_air_2018}. Air quality forecasting systems
allow for sending out warnings of upcoming high pollution episodes to the
population in the short-term, so that appropriate measures can be taken to
minimize as far as possible the damage caused by these episodes. As an example,
the city of Madrid, in order to comply with European regulations
\cite{union2008directive}, devised an air quality protocol which includes
restrictions to the use of polluting vehicles when the concentrations of \no
reach certain thresholds. Consequently, foreseeing the activation of such
restrictions is critical both for the decision makers (which need to announce
them in advance) and for the vehicle owners (which need to plan their transport
alternatives).

The use of data-driven approaches to predict and control air quality is not new.
Following the discussion started by Breiman \cite{breiman2001}, when approaching
a modeling problem two families of methodologies (or ``cultures'', in Breiman
terms) coexist: the data modeling culture, based on the search for a
stochastic data model (for example, time difference equations, in the case of
air quality) that capture the inner behavior of the intervening physical
processes, and the algorithmic modeling culture, based on the use of algorithms
to directly learn the model from data. Given that pollutant concentrations can
be seen as time series, the stochastic data modeling usually deals with using
ARMA based methods \cite{kumar_arima_2010, hassanzadeh_statistical_2009}.
However, this kind of models have trouble handling high non-linearities and high
dimensional environments. To solve this problem, and thanks to the big amount of
data that is gathered nowadays, machine learning models have been applied to
environmental modeling with some success \cite{grivas_artificial_2006,
  navares_predicting_2020}. In this work, however, we advocate for a hybrid
``culture'', in which stochastic data models are combined with algorithmic ones
in a way which permits harvesting the benefits of both approaches while reducing
their disadvantages, as we will show.

Due to the increase in the available computing power and the advances in the
field of neural network-based models, it is nowadays common to find real
applications in which algorithmic modeling is put into practice to predict air
quality. For example, in \cite{nebenzal_long-term_2018} a system is deployed in
which pollution levels based on a threshold are used to study transitions among
states, which makes possible to estimate high pollution episodes in the long
term. In \cite{thatcher_customisable_2010}, authors have combined the TAPM and
CCAM atmospheric models to form a customizable, local-scale meteorological and
air pollution forecasting system, showing that using macroscopic models in the
local scale can provide positive points in the prediction.
\cite{maciag_air_2019} is an example of an ensemble model with a neural network
and an ARIMA, in a similar vein to what will be our proposal, applied
successfully to a real urban environment in the city of London.

For this kind of real applications, it is useful to produce, instead of a
forecast of the expected value of the magnitude under study, an estimation of
the full future distribution, which in turn allows for decision making based on
the probability of the surpassing of certain thresholds. This idea, which will
ultimately be the main goal of the system described below, is common in other
fields and was introduced to air pollution forecasting by
\cite{aznarte_probabilistic_2017}.

The integration of meteorological information and human activities have been
addressed by multiple studies. Some relevant variables, as temperature,
precipitation, or wind speed have shown to be good indicators of pollution
levels \cite{kalisa_temperature_2018, ouyang_washing_2015, kim_influence_2015}.
Also, the physical-chemical mechanisms governing the relationship between these
and air quality has been studied. In \cite{vega_garcia_shapley_2020},
interpretability techniques for deep learning are used to gain insight into
feature importance in a highly similar environment and methodology to ours,
concluding that weather variables, in general, have a high impact when using
machine learning methods for predicting pollution.

However, while these issues have been widely studied from the univariate time
series perspective, the observed spatial interactions between nearby observation
stations might be of importance too as air quality at different stations might
be implicitly related. Spatial-based approaches usually imply assuming or
learning these interdependencies based for example on closeness, but, as it has
been shown \cite{de_medrano_inclusion_2020}, this is not necessarily the most
natural and optimal way to go.

In this paper, we introduce SOCAIRE, the new official air quality monitoring
system for the city of Madrid. This tool, in operational use nowadays, makes use
of both external and internal variables related to air quality in order to
forecast pollutant concentrations. It is a complex modular mathematical system
composed of an ensemble of data manipulation techniques and models that let us
exploit different knowledge in each module: from data cleaning and imputation,
through handling spatial and meteorological non-linear features, to integrating
human behavior and its patterns. By correctly treating all this information, it
is possible to avoid redundancies and to achieve very high performance. As one of
the biggest and most populated cities in Europe, Madrid is a perfect setting
for developing and testing these kind of systems.

The rest of the paper is organized as follows: in Section \ref{S2} the problem
is stated and Madrid's air quality protocol for \no is described, while Section \ref{S3}
presents an analysis and explanation of the different data sources and the data
wrangling process. Section \ref{S4} presents the proposed approach for air
quality forecasting. Then, in Section \ref{S5} we introduce the Bayesian
probabilistic framework that let us accommodate SOCAIRE to the \no protocol. Section
\ref{S6} shows the evaluation of the proposed architecture after appropriate
experimentation and its comparison with other methodologies. Finally, in Section
\ref{S7} we point out future research directions and conclusions.

\section{Problem statement}
\label{S2}

\subsection{Study area and general information}
\label{S2.1}

\begin{figure*}[tbp]
\centering
    \begin{subfigure}{0.4\textwidth}
        \includegraphics[width=\textwidth]{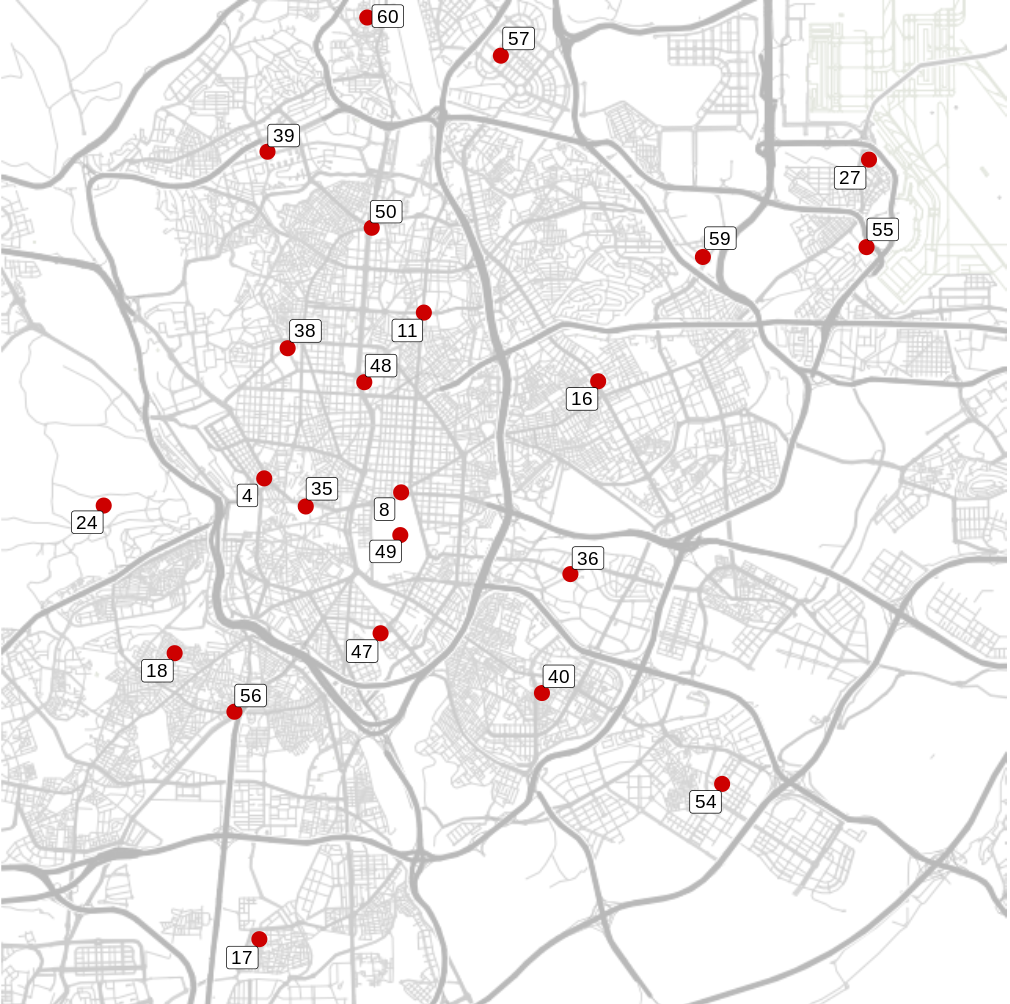}
        \caption{Location of all measurement stations.}
        \label{fig:stations}
    \end{subfigure}
    \begin{subfigure}{0.35\textwidth}
        \includegraphics[width=\textwidth]{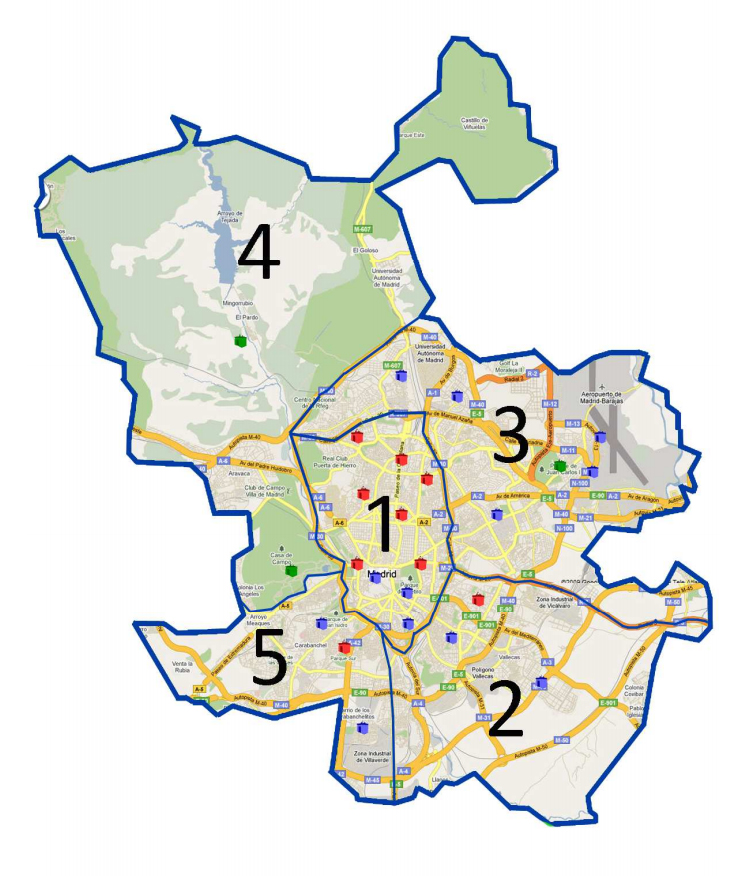}
        \caption{Zones definition for \no protocol.}
        \label{fig:zones}
    \end{subfigure}
    \caption{Location of pollutant measurement stations and definition of the 5
      different areas in which the \no protocol divides the city.}
    \label{fig:maps}
\end{figure*}

\begin{table}[btp]
\centering
\caption{Locations and availability of pollution variables for each station. A
  \checkmark reflects
the presence of data in the corresponding location and for the corresponding pollutant.}
\label{tab:stations}
\scalebox{0.85}{
\begin{tabular}{llllllll}
  \toprule
  Station & Code & Long. & Lat. & NO2 & O3 & PM10 & PM2.5  \\
  \midrule
  Pza. de España & 4 & -3.712 & 40.423 & \checkmark & - & - & -  \\
  Escuelas Aguirre & 8 & -3.682 & 40.421 & \checkmark & \checkmark & \checkmark & \checkmark\\
  Avda. Ramón y Cajal & 11 & -3.677 & 40.451 & \checkmark & - & - & - \\
  Arturo Soria & 16 & -3.639 & 40.440 & \checkmark & \checkmark & - & -  \\
  Villaverde & 17 & -3.713 & 40.347 & \checkmark & \checkmark & - & - \\
  Farolillo & 18 & -3.731 & 40.395 & \checkmark & \checkmark & \checkmark & - \\
  Casa de Campo & 24 & -3.747 & 40.419 & \checkmark & \checkmark & \checkmark & \checkmark  \\
  Barajas Pueblo & 27 & -3.580 & 40.477 & \checkmark & \checkmark & - & - \\
  Pza. del Carmen & 35 & -3.703 & 40.419 & \checkmark & \checkmark & - & - \\
  Moratalaz & 36 & -3.645 & 40.408 & \checkmark & - & \checkmark & - \\
  Cuatro Caminos & 38 & -3.707 & 40.446 & \checkmark & - & \checkmark & \checkmark \\
  Barrio del Pilar & 39 & -3.711 & 40.478 & \checkmark & \checkmark & - & - \\
  Vallecas & 40 & -3.652 & 40.388 & \checkmark & - & \checkmark & - \\
  Mendez Alvaro & 47 & -3.687 & 40.398 & \checkmark & - & \checkmark & \checkmark \\
  Castellana & 48 & -3.690 & 40.439 & \checkmark & - & \checkmark & \checkmark \\
  Parque del Retiro & 49 & -3.683 & 40.414 & \checkmark & \checkmark & - & - \\
  Plaza Castilla & 50 & -3.689 & 40.466 & \checkmark & - & \checkmark & \checkmark \\
  Ensanche de Vallecas & 54 & -3.612 & 40.373 & \checkmark & \checkmark & - & - \\
  Urb. Embajada & 55 & -3.581 & 40.462 & \checkmark & - & \checkmark & - \\
  Pza. Elíptica & 56 & -3.719 & 40.385 & \checkmark & \checkmark & \checkmark & \checkmark \\
  Sanchinarro & 57 & -3.661 & 40.494 & \checkmark & - & \checkmark & - \\
  El Pardo & 58 & -3.775 & 40.518 & \checkmark & \checkmark & - & - \\
  Juan Carlos I & 59 & -3.616 & 40.461 & \checkmark & \checkmark & - & - \\
  Tres Olivos & 60 & -3.689 & 40.501 & \checkmark & \checkmark & \checkmark & - \\
   \bottomrule
\end{tabular}
}
\end{table}

Through this work, we look for a system which is able to predict up to 48 hours
of four of the main existing pollutants: Nitrogen dioxide (\no), ozone
(O\textsubscript{3}), and particulate matter PM10 and PM2.5, where 10 and 2.5
denote the maximum diameters (in micrometers) of the particles. This estimation
needs to be done in the 24 stations that compose the pollution measurement
network, each one with different pollutants. Since one of the main objectives of
the system is to anticipate the activation of mobility restrictions in face of
high pollution episodes, we forecast the main quantiles of the distribution, so
it is easier to make decisions based on pollution level probabilities. Thanks to
its Bayesian estimation of compound events, SOCAIRE becomes an ideal tool to
foresee the scenarios of Madrid's \no protocol, which will be explained
later in this section.

SOCAIRE operates daily on a 48-hours basis: it produces forecasts from 10:00 of
the present day to 09:00 two days later. In the spatial dimension, the
measurement stations of the city council are used as reference points.
Specifically, there are 24 stations distributed throughout the city with sensors
capable of recording different pollutants. Fig. \ref{fig:stations} shows
graphically the location of all the stations. At the same time, the city
considers 5 different areas in the city that are related to the activation of
the \no protocol.
These areas are shown in Fig. \ref{fig:zones}. Table
\ref{tab:stations} shows the correspondence between the different stations and
their code, their location, and the pollutants measured at each one.


\subsection{The \no protocol of the city of Madrid}
\label{S2.2}

In 2018, the city council of Madrid approved an ``Action Protocol for \no
Pollution Episodes'' \cite{noauthor_protocolo_nodate}
(from this point, referred to as ``the \no protocol'') which defines a set of
increasing alert levels, thus classifying the situations of high concentrations
of \no as follows:
\begin{enumerate}
\item PREWARNING: when any two stations in the same area simultaneously exceed
  180 $\mathrm{\mu gm^{-3}}$ for two consecutive hours, or any three stations in the
  surveillance network simultaneously exceed the same level for three
  consecutive hours.
\item WARNING: when any two stations in the same area exceed 200 $\mathrm{\mu gm^{-3}}$
  during two consecutive hours, or any three stations in the surveillance
  network exceed the same level simultaneously during three consecutive hours.
\item ALERT: when in any three stations of the same zone (or two if it is zone
  4) is exceeded simultaneously, 400 $\mathrm{\mu gm^{-3}}$ during three consecutive
  hours.
\end{enumerate}

Depending on the level and the meteorological prospect, a set of increasingly
restrictive mobility limitations will be imposed city-wide by the council with
the aim of mitigating and reducing the negative effects of contamination on the
health and integrity of the population. Thus, the main objective is to know when
and how the conditions leading to the different alert levels will be met, in
order to enable the anticipation of the measures.

\subsection{Framework overview}
\label{S2.3}

\begin{figure*}[tbp]
  \centering
  \includegraphics[width=0.95\textwidth]{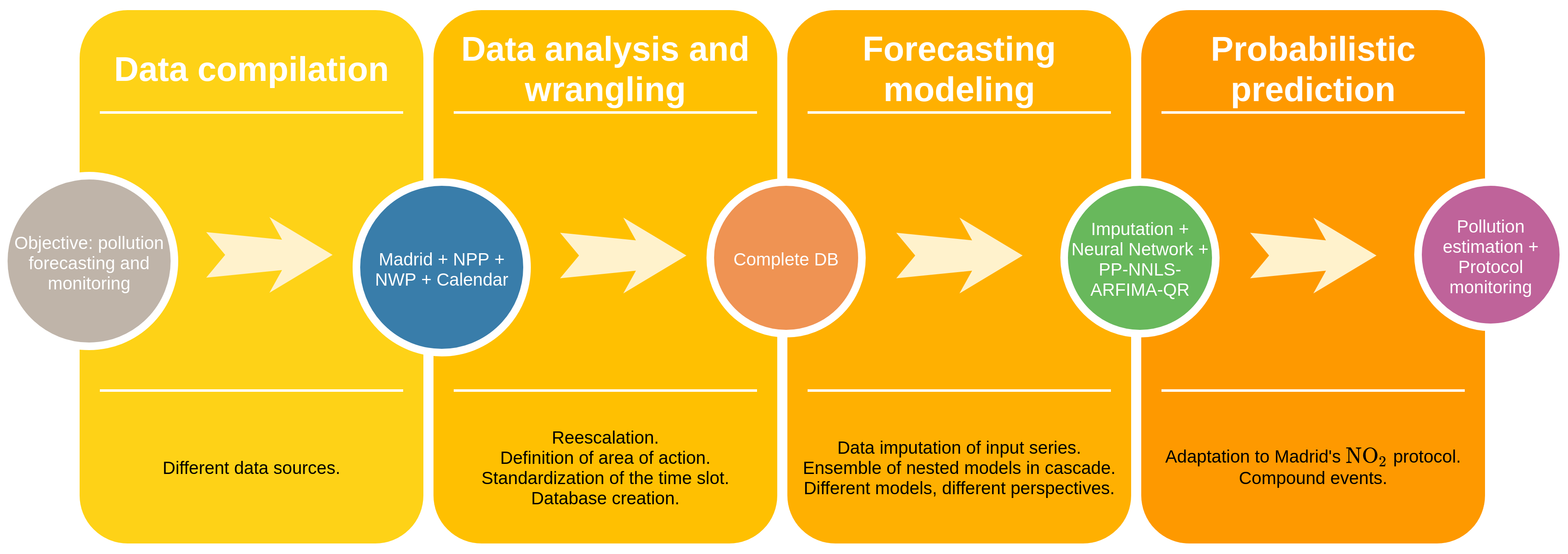}
  \caption{The mathematical components of SOCAIRE.}
  \label{fig:gen_ove}
\end{figure*}

Fig. \ref{fig:gen_ove} presents a summary of SOCAIRE's mathematical structure.
Created to forecast and monitor pollution levels, its operation is based on the
compilation of several data sources which will be described in Section \ref{S3}.
After a proper analysis and cleaning process, the complete database will be used
through an ensemble model composed of a cascade of nested models, each one in charge
of modeling different processes that alter air quality dynamics (Section \ref{S4}). 
Finally, and thanks to the probabilistic nature of the predictions, the system is able to
estimate probabilities from compound events using a Bayesian approach explained
in Section \ref{S5} that is adapted to the aforementioned \no protocol.

\section{Data analysis and wrangling}
\label{S3}

As stated above, in order to aim for the highest performance, SOCAIRE makes use
of all the available information related to the problem. Thus, before
introducing the actual modeling, it is important to present and analyze the set
of available data sources. Concretely, as anticipated, SOCAIRE uses the data of
the concentrations of the different pollutants in the different stations in
Madrid as dependent variables (output) and, as independent variables (inputs),
past pollutant concentrations, numerical pollution predictions coming from the
European CAMS model \cite{noauthor_copernicus_nodate}, numerical weather
predictions served by AEMet \cite{meteorologia_agencia_nodate}, and
anthropogenic information encoding different events such as holidays and school
calendar. The following subsections will detail the origin, peculiarities, and
processing of these data.

\subsection{Pollutants}
\label{S3.1}

\begin{figure*}[tbp]
\centering
\includegraphics[width=1\textwidth]{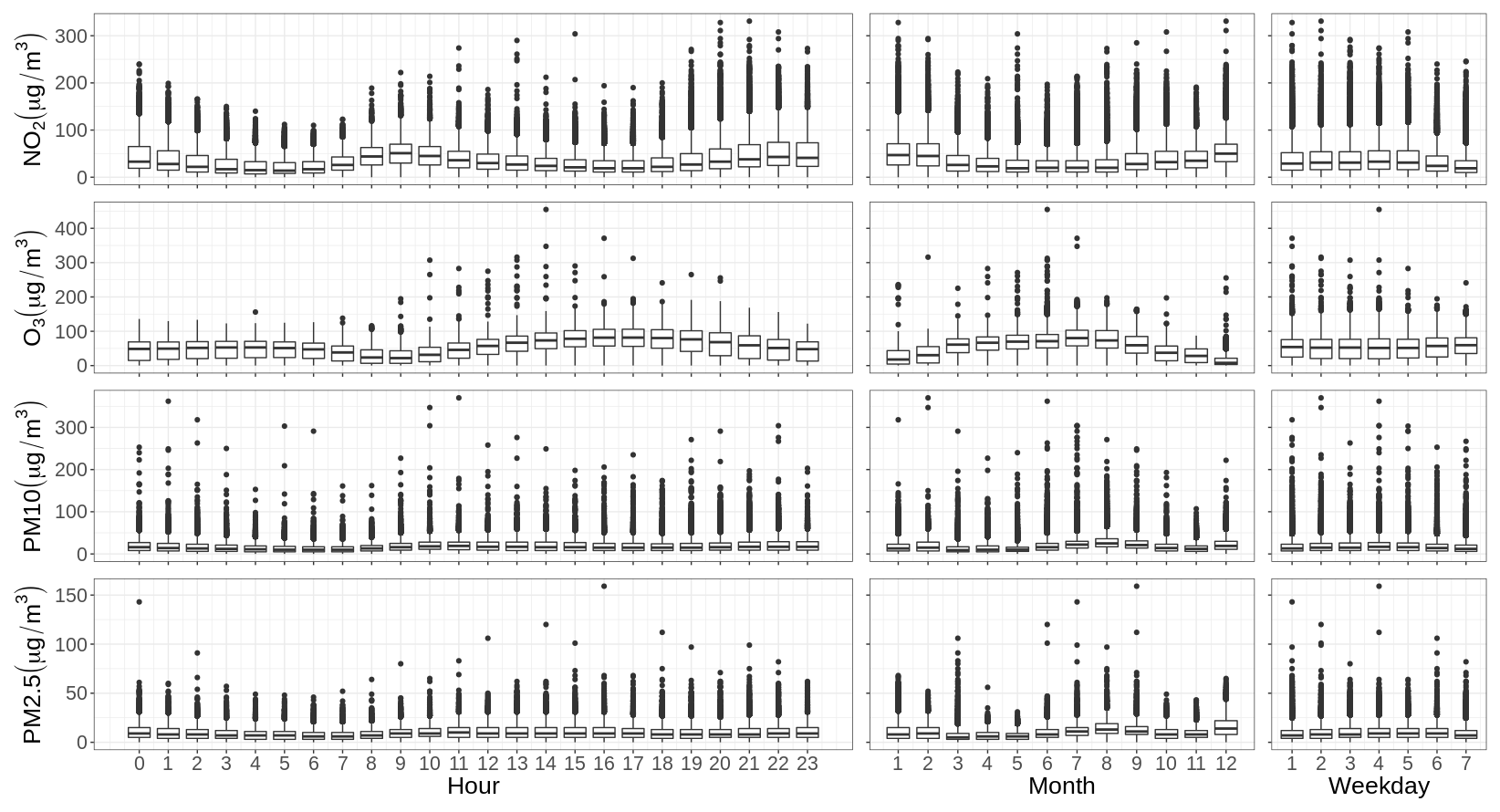}
\caption{Hourly, daily, and monthly temporal distribution of the four target pollutants.}
\label{fig:time_dep}
\end{figure*}

\begin{figure*}[tbp]
\centering
\includegraphics[width=0.8\textwidth]{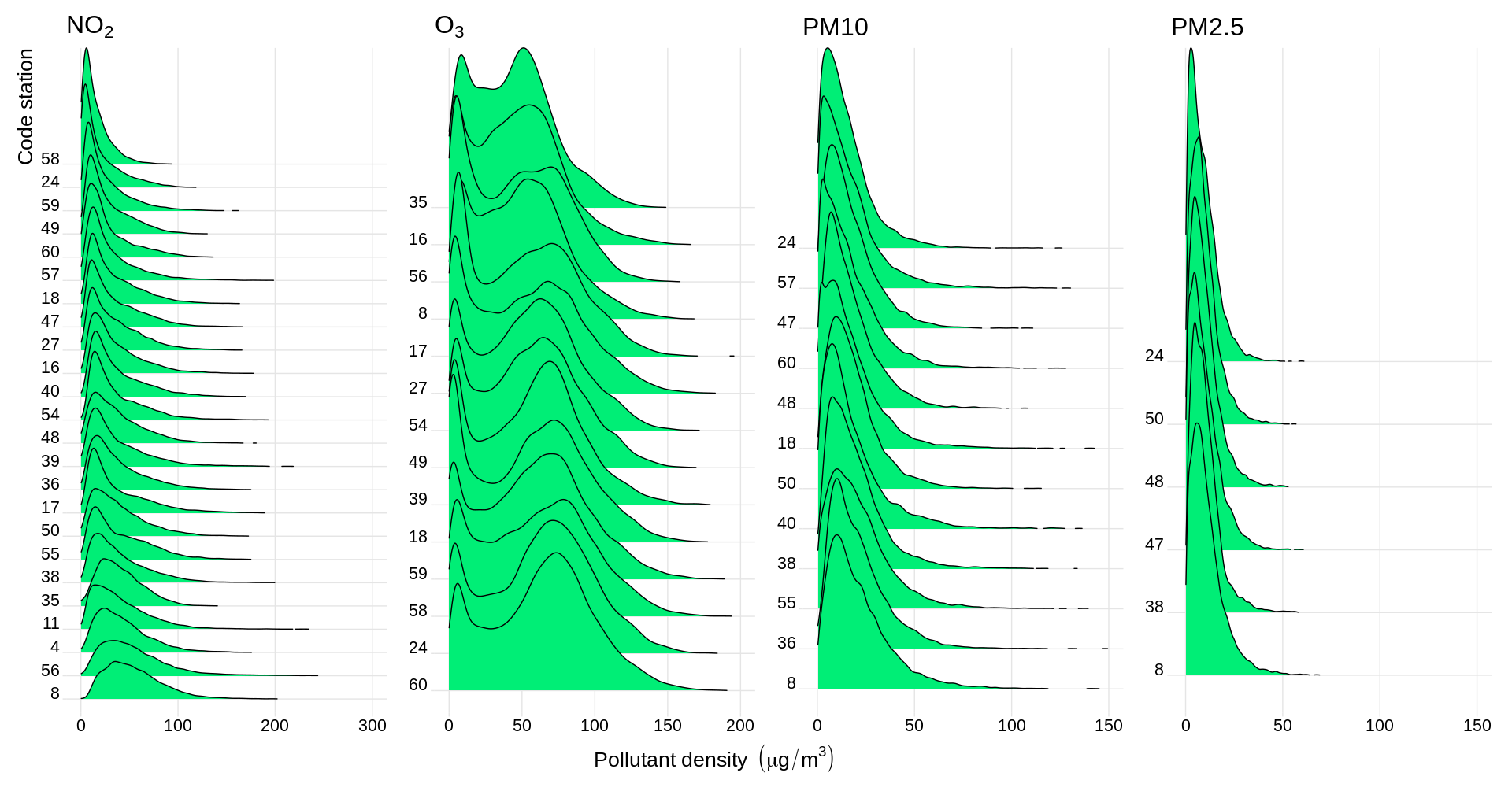}
\caption{Distribution of the series by pollutant and station.}
\label{fig:sp_dep}
\end{figure*}

The temporal behavior of each pollutant series is shown in Fig.
\ref{fig:time_dep}. The daily cycle of all pollutants is dominated in one way or
another by the peak hours of road traffic. Except for ozone, the other three
pollutants to be analyzed have their daily peaks after peak traffic hours. The
\no has the most intense traffic-sensitive cycle, followed by the 10 and 2.5
microparticles, which show a delay of about an hour with respect to the \no.
O\textsubscript{3} presents a daily cycle that is practically inverted with respect to the
rest.

Everything said for the daily cycle applies to the weekly cycle, with weekend
being days with lower levels of traffic. It can be assumed that holidays and
long weekends will behave as public holidays, so the forecast model would have
to take this into account. As expected, the daily cycle is not independent of
the weekly one, but each day of the week has its own cycle, especially different
on weekends from working days.

In the annual cycles, a greater variety of behaviors can be observed. All
pollutants, especially ozone, rebound in summer except \no which has the
opposite behavior in this case.

Respect to the spatial dimension, Fig. \ref{fig:sp_dep} represents the
empirical distributions for each pollutant. It can be seen that all stations
report a similar behavior, without clear relation patterns between closeness
and distribution. This fact will be of interest later when taking into account
these spatial relationships in the modeling process.

As the distributions show a clear asymmetry, logarithmic transformations are
used. Pollutant data is publicly available at the \textit{Open data portal of Madrid}
\cite{noauthor_en_nodate}.

\subsection{Numerical weather predictions (NWP)}
\label{S3.2}

As mentioned in Section \ref{S1}, meteorology has shown to be especially
important for air quality. Hence, having weather forecasts for the period in
which the air quality forecasting is being made is expected to positively impact
the precision of the forecasts. In this work, we use NWP from the Integrated
Forecasting System (IFS) of the ECMFW \cite{blanchonnet_set_2015}, for the
following set of variables:
\begin{itemize}
\item \textbf{Boundary layer height} (in meters): This parameter is the depth of
  air next to the Earth's surface which is most affected by the resistance to
  the transfer of momentum, heat or moisture across the surface. The boundary
  layer height can be as low as a few tens of meters, such as in cooling air at
  night, or as high as several kilometers over the desert in the middle of a hot
  sunny day. When the boundary layer height is low, higher concentrations of
  pollutants (emitted from the Earth's surface) are found.
\item \textbf{Surface pressure} (in Pa): This parameter is the pressure (force
  per unit area) of the atmosphere on the surface of land, sea, and in-land
  water. It is a measure of the weight of all the air in a column vertically
  above the area of the Earth's surface represented at a fixed point. Air
  pollution is especially prominent where high pressure dominates. Subsiding
  motions within an anticyclone suppress air trying to rise off the surface.
  Adiabatic warming of subsiding air creates a subsidence inversion which acts
  as a cap to upwardly moving air. Pollution problems dissipate when a low
  pressure system replaces a retreating anticyclone.
\item \textbf{Temperature} (in K): This parameter is the temperature of air at
  2 m above the surface of land, sea, or in-land waters. Generally, higher
  temperatures and hotwaves are directly related to episodes of higher
  pollution levels.
\item \textbf{Precipitation} (in mm): This parameter is the accumulated liquid
  and frozen water, including rain and snow, that falls to the Earth's surface.
  It is the sum of large-scale precipitation (that precipitation which is
  generated by large-scale weather patterns, such as troughs and cold fronts)
  and convective precipitation (generated by convection which occurs when air at
  lower levels in the atmosphere is warmer and less dense than the air above, so
  it rises). Precipitation parameters do not include fog, dew, or the
  precipitation that evaporates in the atmosphere before it lands at the surface
  of the Earth. Air pollution is typically negatively correlated to the quantity
  of rainfall, existing a so called washing effect of precipitation.
\item \textbf{U wind component} (in $\mathrm{ms^{-1}}$): This parameter is the
  eastward component of the 10m wind. It is the horizontal speed of air moving
  towards the east, at a height of ten meters above the surface of the Earth.
  Pollutants tend to pile up in calm conditions, when wind speeds are not more
  than about 3 $\mathrm{ms^{-1}}$. Speeds of 4 $\mathrm{ms^{-1}}$ or more favour
  dispersal of pollutants, which, literally, clears the air.
\item \textbf{V wind component} (in $\mathrm{ms^{-1}}$): This parameter is the northward
  component of the 10 m wind. It is the vertical speed of air moving towards
  the north, at a height of ten meters above the
  surface of the Earth. Again, wind is highly related to pollution dissemination. 
\end{itemize}

NWP are interpolated to the location of each station of the air quality
monitoring network. As pointed out previously, these forecasts are provided by
AEmet in an hourly basis.

\subsection{Numerical pollution predictions (NPP)}
\label{S3.3}

CAMS (Copernicus Atmosphere Monitoring Service)
\cite{noauthor_copernicus_nodate} provides a four day-horizon hourly pollution
forecast which covers all Europe on a synoptic scale. The model takes into
account global and regional numerical weather predictions from the ECMWF
\cite{marecal_regional_2015}, as well as other types of forecasts about the
production of certain chemicals of natural and human origin from models such as
C-IFS Forecasts or CAMS 81.

All these models always refer to a geodesic grid of between 10 and 20 km on each
side, so it is not very sensible to use them to directly forecast the
concentrations with the resolution required inside a city, which might well be
below one kilometer.


\subsection{Anthropogenic features}
\label{S3.4}

As we saw in Figure \ref{fig:time_dep}, depending on the human activity the
temporal patterns of the series are different. Similar to weekends and months,
public holidays and other designated days, as well as the school calendar, have
a significant influence on road traffic, giving rise to a very different daily
cycle. In special dates, we usually find a lower intensity in the center but a
punctual growth in other places, particularly on the main access roads to the city
related to holiday departures and returns.

Also, each type of calendar effect has different effects on each hour of the
day. In addition, some of them can fall on Saturday or even on Sunday, in the
case of Christmas Eve and New Year's Eve, and it is clear that the effect cannot
be the same as when it falls during the week, so all these issues must be taken
into consideration.

In our particular case, we will take into account the following aspects:
\begin{itemize}
\item \textbf{Public holidays}: Public holidays, long weekends, and special
  days, such as Christmas Eve and New Year's Eve, are characterized by
  significantly less road traffic than a normal working day (apart from other
  departure and return operations that may occur on some of these days and which
  will be taken into account later).

  It has been observed that public holidays have different effects, both in
  terms of level and intraday evolution, depending on their location within the
  year, probably due to climate reasons, hours of light, and living patterns.

\item \textbf{Holiday departures and returns}: Extraordinary periods such as
  bank holidays, long weekends, or even weekends cause a temporary exodus of
  citizens with large accumulations of vehicles in the so-called departure and
  return operations.

  Departure operations can take place during the evening of the eve of the first
  non-working day or during the morning of that day, while return operations
  occur mostly during the evening of the last holiday, sometimes reaching the
  early morning of the next working day.

  As with other variables, the effect varies with the hours within a relatively
  soft form.
    
\item \textbf{School Calendar:} in Spain, school calendar and schedule is highly
  related to usual hourly, weekly, and monthly patterns and so, it can model
  with high precision the daily living. The school day can be complete or
  normal, average (pre- and post-holidays) or non-existent, either in isolation
  or for summer, winter nor spring holidays. Each type of day other than the
  normal one is introduced as an effect with a different intraday cycle between
  07:00 and 08:00. 
\end{itemize}

By combining all these variables, we ensure that the information relating to
human mobility in the city is covered, both for normal situations and for
special events. These exogenous variables are defined for each station, as not
all parts of the city have the same dynamics.

\subsection{Data wrangling}
\label{S3.5}

When working with such diverse data sources, is usual to deal with very
heterogeneous formats and criteria, which implies that pre-processing and
cleaning steps are of utmost importance. Some of the most important ones for
this project are listed in this section.

Firstly, some sources use UTC time and others use Madrid's local time. In
addition, the processes that transfer data between different programming
environments (R, TOL, and Python) also have to take into account that each of
these systems work differently with respect to winter and summer daily savings
time changes.

Secondly, both NWP and NPP distribute their forecasts in a different geodesic
grid, which in turn does not coincide with the coordinates of the pollution
monitoring stations. At first, an attempt of interpolation was made by using the
three closest grid points to each station as drivers, but it soon became
apparent that this was an excessive complication with very little added value,
as the forecasts were highly correlated. Therefore, in the final version, only
the nearest reticular point to each station is used.

Thirdly, weather predictions are not always in the most appropriate metric,
so it is necessary to create derived variables that serve better as drivers of
the models. To begin with, there are variables that change scale throughout
history and it is necessary to unify the criterion for obtaining uniform series
in time. Then, there are other variables that are interesting to modify
conceptually, for example, instead of the east-west and north-south coordinates
of wind speed, it is much better to use scalar speed, which is the fundamental
factor of diffusion, and direction, which is less important. Finally, it is
known that meteorological factors not only have an instantaneous effect, but
also a delayed effect that can be exercised up to a few hours later. For this
reason, some variables delayed up to 4 hours have been created and integrated
with the rest of features.

Finally, since we are dealing with a cascade-like ensemble of models, in which
the output of one is the input of another which may require a substantially
different structure, each level of modeling requires a series of steps to
prepare the data to be as expected in the next phase.

Let us note that the most laborious part of the data pre-processing has been the
imputation of missing values. However, given the importance of this part, it has
been decided to include imputation of data as part of the modeling strategy and
is explained later in section \ref{S4.1}.

\section{Modeling Strategy}
\label{S4}

The concentration of a given pollutant in the air depends on at least two
conceptually distinct groups of factors:
\begin{itemize}
\item \textbf{Emission factors}: generally these are of a social order, such
  as road traffic or heating, which are predictable to some extent, although
  there are also totally unpredictable events such as fires, and others that
  could be anticipated to some extent such as strikes or sporting events with a
  multitudinous following.
\item \textbf{Dispersion factors}: basically these are consequences of the
  weather conditions on which there are quite precise forecasts on the horizon
  of 2 or 3 days ahead.
\end{itemize}

Note that a certain factor, such as rain, can work in both directions at the
same time: on the one hand it can cause an increase in traffic on a normal
working day, which increases pollution, but on the other hand it disperses,
especially the particles as they are carried to the ground, which decreases
pollution. It is even possible that the effect is different depending on the
day and time. Following the example of the rain that normally increases the
traffic in a working day, it can on the contrary contract the traffic in an exit
operation, when it will discourage people to leave the city.

This causal complexity, added to the high degree of interaction between factors,
makes the phenomenon highly unstable and therefore very difficult to predict
using any individual methodology. For this reason, an ensemble model composed
of a cascade of nested models
has been designed, such that the output of each is used in the next to
get the most out of each:
\begin{itemize}
\item \textbf{Imputation techniques}: Although this task is usually framed as
  part of the data wrangling process, in this project it involves the
  development of models of some complexity, due to the fact that the omitted
  elements are presented with a certain frequency and not always in a sporadic
  way, but covering periods of time that can even be of several weeks. These
  techniques are detailed in Section \ref{S4.1}.
\item \textbf{NNED model}: a special flavor of convolutional neural networks
  called neural net encoder-decoder, which, using as inputs the outputs of the
  imputation models, allows to jointly forecast the concentrations of a
  pollutant in all the stations at the same time. It takes into account the NWP
  and NPP, as well as the recent past of all stations for each input variable,
  including the previous pollution itself, and is capable of automatically
  detecting non-linearities and interactions between different features.
  However, it does not allow for the natural treatment of irregularities in
  non-cyclical anthropogenic factors related with traffic. It is described in
  detail in Section \ref{S4.2}.
\item \textbf{PP-NNLS-ARFIMA-QR model}: This is a chain of models by itself
  developed specifically to deal with anthropogenic factors in a Bayesian way.
  It will be explained in detail in Section \ref{S4.3}.
\end{itemize}

\subsection{Imputation techniques}
\label{S4.1}


\begin{figure*}[tbp]
    \centering
    \includegraphics[width=0.95\textwidth]{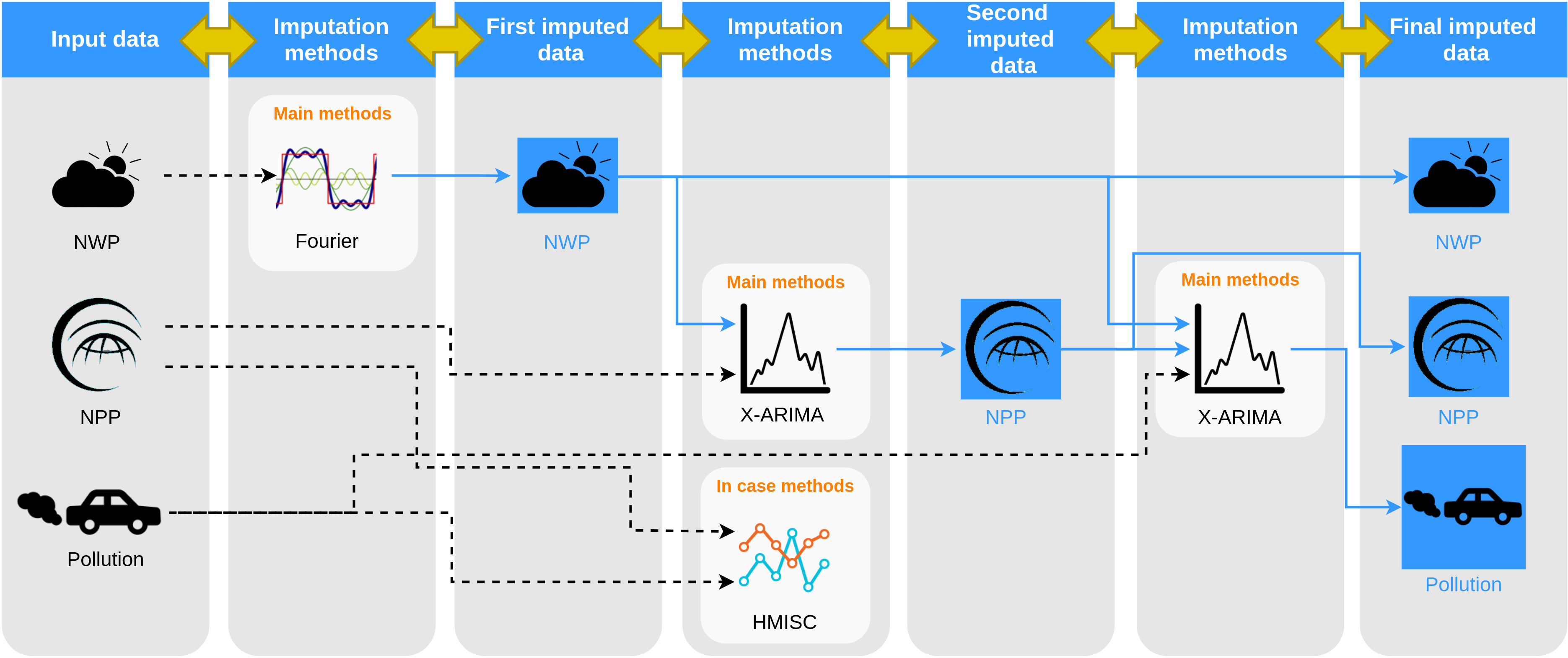}
    \caption{Imputation scheme. All data sources likely to have missing
      values are part of the process. Through successive refinements, the
      initial data are processed through different techniques in a way that
      takes into account the nature and particularity of each source when
      performing the task. Black-dashed arrows follow processes in which data
      has not been imputed yet, while blue-solid arrows formalize the idea that
      this particular set has already been imputed.
      }
    \label{fig:imp}
    \end{figure*}

In the different data sources, it is relatively frequent to find missing data
that can cause problems in the modeling process. For this reason, it is
necessary to devise a sensible way to fill in these missing values, replacing
them with approximate or expected values by a series of auxiliary models. When
there are only very sporadic omissions of short duration, it might be sufficient
to apply some kind of approximation by interpolation, but there might be up to
consecutive weeks of data omitted in several or all variables from one or more
sources at the same time. Thus, in order to develop a robust operational system,
able to function even in the presence of missing data, more complex and specialized
techniques are required. 

\subsubsection{Trigonometric interpolation}
\label{S4.1.1}

First, a trigonometric interpolation is used as a univariate method to generate
sensible values for those series with clear cyclical components, such as
temperature. In our case, these series present very few omissions, so we
consider this technique to be sufficient. Since the data are arranged in a
regular grid, this can be done by the discrete Fourier transform.
 
\subsubsection{Multiple Imputation using Additive Regression, Bootstrapping, and
  Predictive Mean Matching (HMISC)}
\label{S4.1.2}

Multiple imputation using additive regression, bootstrapping, and predictive
mean matching consists of drawing a sample with replacement from the real series
where the target variable is observed (i.e. not missing); fitting a flexible
additive model to predict this target variable while finding the optimum
transformation of it; using this fitted model to estimate the target variable in
all of the original series; and finally, imputing missing values of the target
with the observed value whose predicted transformed value is closest to the
predicted transformed value of the missing value. This methodology is
implemented in the R package \textit{HMISC} \cite{jr_hmisc_2020}. As the
meteorological variables have already been imputed with the previous method
(which will be used as input here), it is only applied to the NPP and the
pollutant concentrations themselves. This method is actually used for safety in
case the next one (X-ARIMA) fails. As several parts of the framework can not
handle missing data, this step is required in order to assure proper
functioning.


\subsubsection{X-ARIMA}
\label{S4.1.3}

Once the previous two standard imputation methods are applied, it is turn for a
univariate dynamic causal imputation method. It analyses how both the present
and the past of a group of variables, including the target variable itself, act
on the future of this target variable. These models are quite complex and, to
improve the imputation, they are applied in two successive phases: in the first
one, the NPP are imputed as a function of the NWP; in the second one, the
pollution observations are imputed as a function of the NWP and the NPP.

Mathematically speaking, we have that, being $Y_t$ the time series of
concentration of the pollutant in question and $X_{t,k}$ the linearized
inputs from the explanatory terms described above, the general formula of the
Box-Jenkins' X-ARIMA models \cite{box-jenkins-1976} used is as follows (where,
as usual, $B$ is the backshift operator):
\begin{equation}
  \Delta\left(B\right)\phi\left(B\right)\left(Y_{t}-\underset{k=1}{\overset{K}{\sum}}X_{t,k}\alpha_{k}\right)=\theta\left(B\right)\epsilon_{t}.
\end{equation}

The summation $\sum_{k=1}^K X_{t,k}\alpha_{k}$ will be called the filter of
exogenous effects while the equations in differences expressed by the delay
polynomials will be called endogenous factors or the ARMA part of the model.

Note that this model is very different from the typical ARIMA model with
exogenous effects of the ARIMA-X class
\begin{equation}
\Delta\left(B\right)\left(\phi\left(B\right)Y_{t}-\underset{k=1}{\overset{K}{\sum}}X_{t,k}\alpha_{k}\right)=\theta\left(B\right)\epsilon_{t},
\end{equation}
which is easier to estimate but also is considered to be much less effective in
explaining the phenomena that actually occur in real life.

\begin{itemize}
\item \textbf{Exogenous factors}: The NWP series has only very few isolated
  omitted data and in principle there is no reason to think that they will occur
  more frequently in the future. For this reason, it is more than sufficient to
  use an imputation system based on the Fourier transform.

  The imputation of the NPP series will take as inputs the previously imputed
  NWP. Specifically, the boundary layer height (BLH), wind speed (WS), and
  precipitation (TP) have been used, applying different Box-Jenkins time
  transfer functions \cite{box-jenkins-1976} with different damping parameters
  in order to collect in a more synthetic way the time delayed transfers already
  discussed.
  For the series of pollution observations, both NWP and NPP will be used, after
  all of them have been already imputed.
\item \textbf{Endogenous factors}: 
  The ARIMA polynomials in this case are multi-seasonal. Among the inertial
  factors of the stochastic process, and besides the regular time (hourly), both
  the daily cycle of periodicity $24$ hours and the weekly cycle of periodicity
  $24 \times 7 = 168$ hours are taken into account.

  Obviously, there is also a pseudo annual cycle and a trend but they will be
  filtered by some of the explanatory drivers or exogenous factors indicated in
  the previous section. On the one hand, the annual cycle is not in harmony with
  the weekly or daily cycle, that is, its periodicity is not a whole number, and
  on the other hand it is enormous: $365.2425 \times 24 = 8765.82$, so it is
  practically intractable for the ARIMA approach in an hourly series. Even in a
  daily series it presents serious difficulties and consumes a lot of resources.
\end{itemize}

A complete overview of the imputation process is shown in Fig. \ref{fig:imp}.

\subsection{Neural network encoder-decoder: NNED model}
\label{S4.2}

\begin{figure*}[tbp]
\centering
\includegraphics[width=0.25\textwidth]{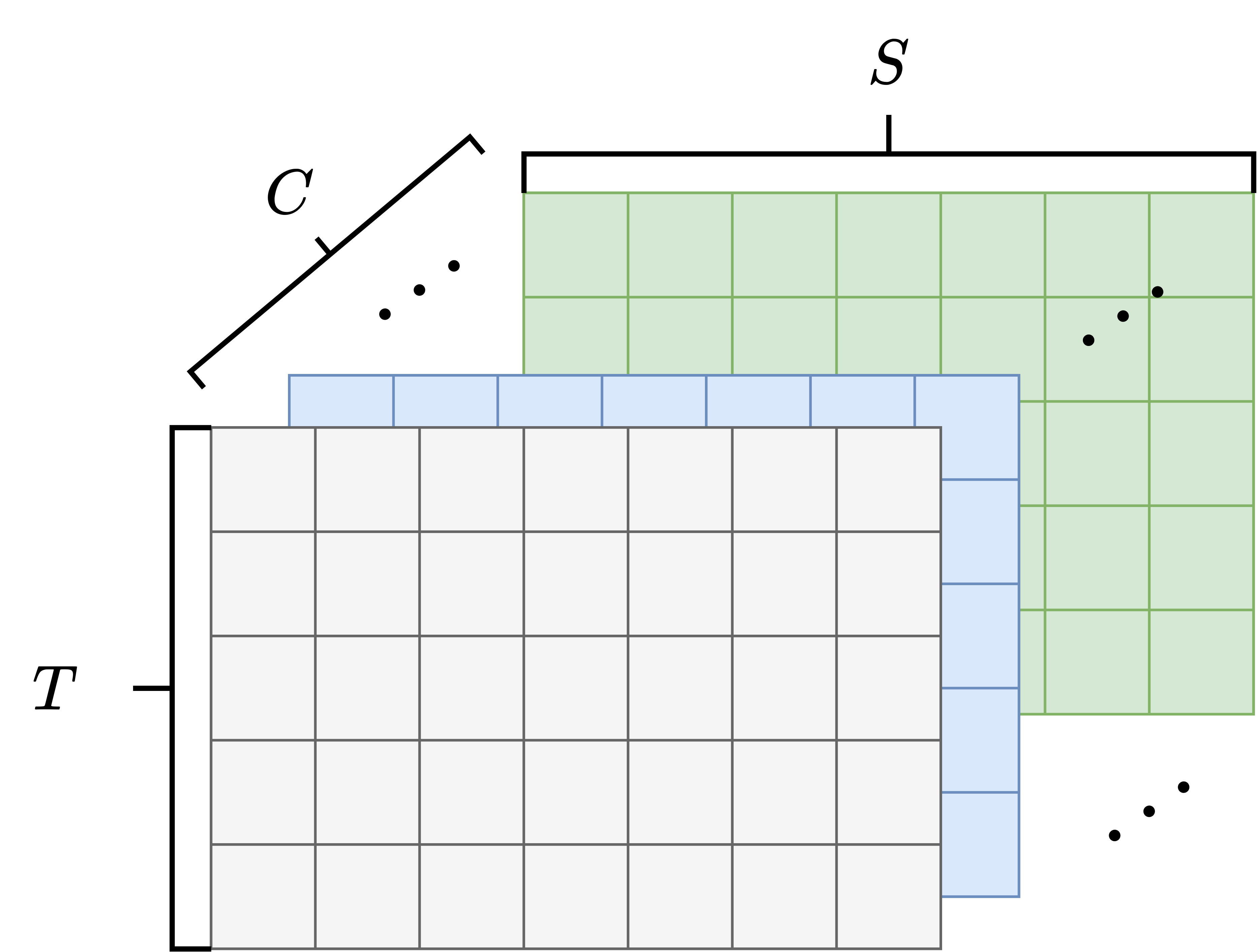}
\caption{Scheme for the input sequence of the NNED model. As long as all
  variables are spatio-temporal and have an equivalent structure for both
  dimensions, these sequences can be easily introduced as $C \times T \times S$
  images, with variable, temporal and spatial dimension respectively.}
\label{fig:sann_1}
\end{figure*}

\begin{figure*}[tbp]
\centering
\includegraphics[width=0.95\textwidth]{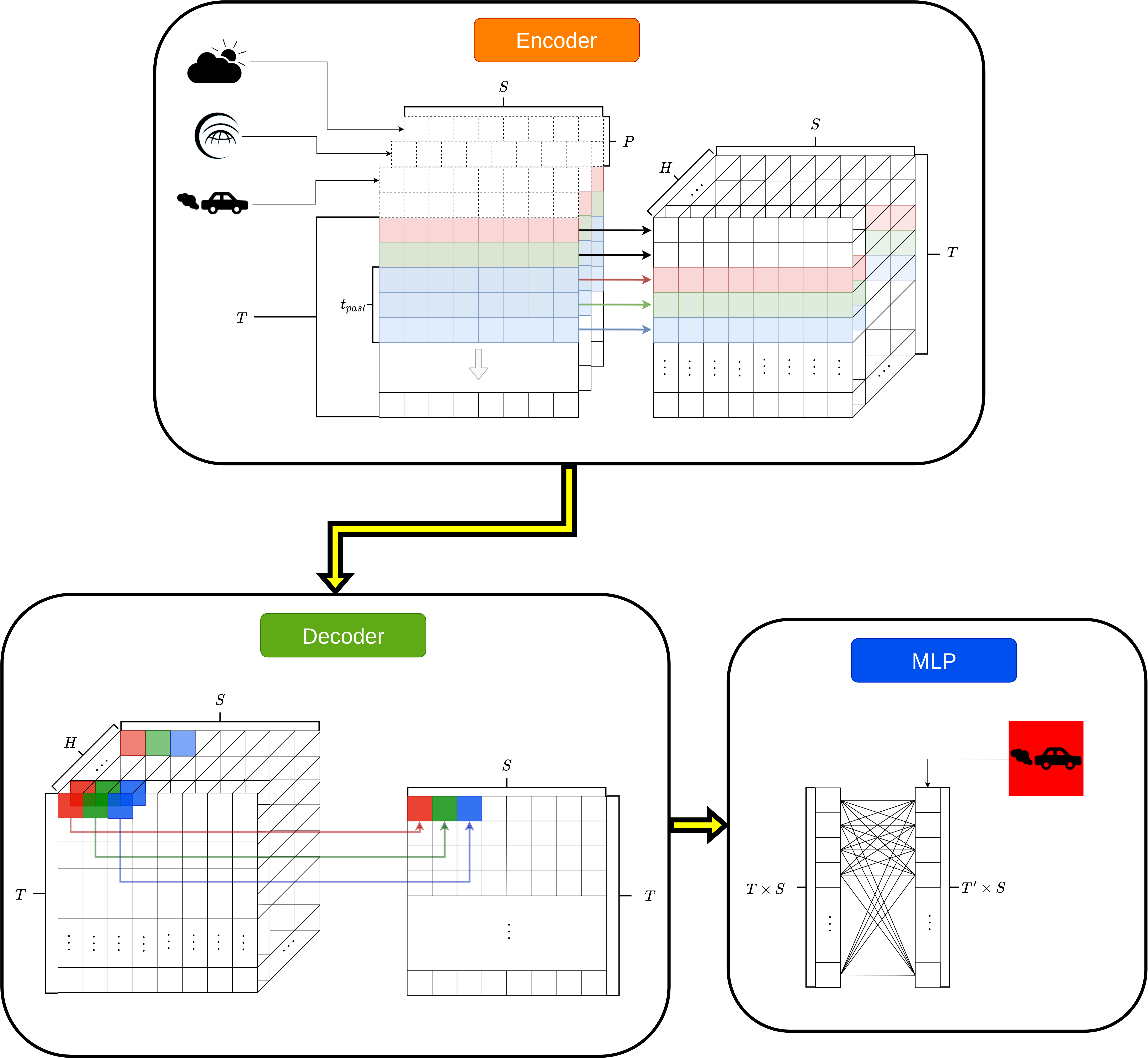}
\caption{NNED scheme. NNED consists of an encoder with agnostic
  convolutions ($k_{1} = t_{\mathrm{past}}$, $k_{2} = S$), a decoder with $1\times1$
  convolutions, and a dense layer that relates all input information with all
  output elements. In red, forecast pollutant concentration. 
  }
\label{fig:nned}
\end{figure*}

Given that interactions between pollution itself and other relevant features, as
NWP, show a complex and highly non-linear behavior in both time and space, deep
learning arises as a suitable mathematical solution. No anthropogenic
interactions are modeled at this point. A step forward with respect to the usual
deep learning architectures, NNED model is based on the idea of spatial
agnosticism for solving spatio-temporal regression problems
\cite{de_medrano_inclusion_2020}.
It has been shown that when the spatial granularity of the series is low and its
spatial autocorrelation is close to 0, traditional convolutional neural networks
(CNN) fail to extract all the information from the series as the adjacency
assumption for learning shared-weights does not entirely hold. That way, it is
possible to obtain better prediction performance by avoiding traditional CNNs by
using a spatially agnostic version of convolution.

By spatial agnostic network, we refer to a neural network in which no spatial
information is introduced and past temporal information can be handled and
introduced in the calculation of each new state. In order to do so, the input
sequence scheme relies upon a $C \times T \times S$ images as shown in Figure
\ref{fig:sann_1}, where the number of channels $C$ represents the number of
input spatio-temporal variables. Similar to the usual input scheme presented in
graph neural networks, this methodology let us treat both spatial and temporal
dimension simultaneously. For our concrete case, the input series will be pollution,
NWP and NPP for all stations during the past 48 hours. The model will output
pollution forecasting for all stations for the next 48 hours.

NNED is composed of three different modules:
\begin{itemize}
\item \textbf{Encoder}: It is in charge of coding the input information of the
  space-time series in a space of superior dimension $H$. That is, it increases
  the expressiveness of the input by relating all the variables to each other.
  As we expect this model to work without spatial information, the encoder
  needs some modifications in its convolution scheme. The convolution itself
  ($*$ operator) has the usual form for 2D images given an input $x$:
  \begin{equation}
    \label{eq:1}
    (x * K)(i,j) = \sum_{m}^{k_{1}}\sum_{n}^{k_{2}} x(m,n)K(i-m, j-n) 
  \end{equation}
  where $K$ is the learnable kernel. However, the kernel size is regularly used
  with equivalent values for its two dimensions $k_{1} = k_{2} = k$. In this
  case, not only this kernel uses different values for each component, but
  kernel size for spatial dimension must be equal to the number of spatial
  zones: $k_{2} = S$. As a result, the convolution operation is made over all
  locations at once. The kernel size in the temporal dimension is defined as
  $t_{\mathrm{past}}$ and needs to be fixed as part of the network architecture.
    
  The temporal dimension is dominated by a causal convolution. Generally, causal
  convolution ensures that the state created at time $t$ derives only from
  inputs from time $t$ to $t - t_{\mathrm{past}}$. In other words, it shifts the
  filter in the right temporal direction. Thus $t_{\mathrm{past}}$ can be
  interpreted as how many lags are been considered when processing an specific
  timestep. Given that previous temporal states are taken into account for each
  step and that parameters are shared all over the convolution, this methodology
  might be seen as some kind of memory mechanism by itself. Unlike memory-based
  RNN (like LSTM and GRU) where the memory mechanism is 
  learned via the hidden state, in this case $t_{\mathrm{past}}$ acts as a
  variable that lets us take some control over this property.
    
  In order to ensure that each input timestep has a corresponding new state when
  convolving, a padding of $P=t_{\mathrm{past}}-1$ at the top of the image is
  required. To guarantee temporal integrity, this padding must be done only at
  the top. By using convolution in this form, once the kernel has moved over the
  entire input image $T \times S$, the output image will be $T \times 1$. Now,
  if we repeat this operation $H \times S$ times, we will create a new hidden
  state with $H$ channels and an output image with $H \times T \times S$
  dimensions.
    
  Thus, we have coded input information relating all variables among them
  without exploiting \textit{prior} spatial information based on adjacency.
    
\item \textbf{Decoder}: Its function is to decode the information contained in
  the hidden space of high dimensionality. To do this, it learns how to merge
  the $H$ hidden states present for each input and location timestep into a
  single value. Because this information is expected to be similar throughout
  the image, a kernel of size $k_1 = k_2 = 1$ is used. Thus, it changes from an
  image $H \times T \times S$ to, again, a $T \times S$.
    
\item \textbf{Multilayer perceptron}: Finally, a multilayer perceptron of input
  $T \times S$, and output $T' \times S$ is used, relating each element
  obtained by the processes of coding and decoding with each of the zones and
  times to be predicted. The output of this multilayer perceptron is the output
  reported by the NNED model.
\end{itemize}

Finally, the complete procedure for this model is described graphically in Fig
\ref{fig:nned}.

\subsection{PP-FSLR-ARFIMA-QR model}
\label{S4.3}

The PP-FSLR-ARFIMA-QR model is actually a chain of models itself, which has
been developed specifically to address the anthropogenic factors that in this case are
of the non-cyclical calendar type. It is true that there is an underlying weekly
cycle, but due to holidays and long weekends, and the interaction with the
annual cycle (a long weekend in spring is not the same as in winter), it
presents strong distortions that have to be dealt with \textit{ad hoc}. Thus,
this model uses the different initial data sources and knowledge learned from
previous modules to exploit all this information in order to return a
probabilistic prediction for the next 48 hours. In this case, a different model
is adjusted for each station.

\subsubsection{PP: daily classification into pseudo-periodic sub-dates}

Principally, the PP module is responsible for dividing the time sequence
according to the type of day, depending on its position at weekends and
holidays:
\begin{itemize}
\item \textbf{Post}: After a long weekend (usually Monday).
\item \textbf{Ext}: Both the day before and the day after are working days
  (usually Tuesday-Thursday).
\item \textbf{Prev}: Weekend or Holiday Eve (usually Friday).
\item \textbf{First}: First day of a long weekend or weekend (Saturday mostly).
\item \textbf{Int}: Internal to a long weekend, excluding the first and last
  day.
\item \textbf{Last}: Last day of a holiday or weekend (Sunday as a rule).
\end{itemize}

For each one of these 6 possibilities, a time series is generated and a chain of
models (described below) is developed.

\subsubsection{FSLR: fixed sign linear regression}

Once the type of day has been determined, we start with a linear regression
whose coefficients are forced to be non-negative based on the work of
\cite{lawson1995solving}.
If a driver should have a negative effect, it is introduced with a change of
sign. This Bayesian approach is not very common, but it is very appropriate in
many occasions, since we often do not have a very detailed quantitative
information about the form of the distribution of the typical prior conjugate
\cite{Fink1997ACO}
, but we do have a very clear qualitative knowledge, for example with respect to
the sign that it should take, which can be expressed as a uniform distribution
in the semimark $x \ge 0$ or $x \le 0$.

The effects considered in this regression are:
\begin{itemize}
\item \textbf{Instantaneous NNED forecast}: The main driver is the forecast made
  with the neural network model explained in Section \ref{S4.2}. In the case of
  the \textbf{Ext} type of day it is diversified according to the day of the
  week which can be Tuesday, Wednesday or Thursday, as it has been observed that
  a certain differences exist. In the rest of sub-dating, the case of days of the
  week does not allow for such diversification.

\item \textbf{Daily inertia (medium term)}: The average of the already known
  observations with 23, 24 and 25 hours of delay on the one hand, and with 47,
  48 and 49 on the other. By forcing the positive sign, the inertia is
  maintained if it is significant and positive. In other cases, the NNED
  algorithm itself is in charge of canceling it. It works approximately as a
  kind of autoregressive seasonal model of period 24 in the natural time dating,
  as opposed to the artificial time division subdate just described in the
  previous section.

\item \textbf{Daily correction (medium term)}: The average of the errors made by
  the model itself with 23, 24, and 25 hours of delay on the one hand and with
  47, 48, and 49 on the other, which are also known. In this case they will be
  used with the opposite sign, that is, if an error is made in one direction it
  is corrected in the other, provided that such effect has been estimated as
  significant, and otherwise the NNED cancels it out. It works approximately
  like a kind of moving average seasonal model of period 24 in natural time
  dating.

\item \textbf{Inertia and time correction (short term)}: For the first hours of
  the morning of each forecast session, the observations and errors of the last
  hours are also available, so it is possible to build inertia and short-term
  correction inputs similar to the two previous ones. From midday of the same
  forecast day they are no longer useful. They work as a kind of regular ARMA
  in natural time dating.

\item \textbf{Protocol Activation}: When the mobility restrictions imposed by
  the \no protocol described in Section \ref{S2.2} are activated, the pollutant
  concentrations might be reduced with greater or lesser success, so that the
  NNED forecasts become obsolete and must be intervened in a deterministic way.
  They are entered with a negative sign because it would not make sense for the
  action to increase contamination

\item \textbf{Workday indicator}: Within a long weekend, pollution is particularly
  reduced on the public holidays themselves, so a slight upward
  correction is needed for the rest of the days of the long weekend. It only
  affects the type of day \textit{Int}.

\item \textbf{School Calendar}: During school vacations and adaptation periods
  with reduced schedules at the beginning and end of the school year, there is a
  certain reduction in pollution that suggests a downward correction.
\end{itemize}

Concretely, this regression is estimated in logarithmic terms of both the
observations and the NNED forecasts and errors, since it has been observed that
the multiplicative relationship predominates over the additive.

\subsubsection{Dynamic Regression ARFIMA}

On the errors of the previous regression, a regular dynamic model is developed
(without a seasonal part) that is concerned with maintaining inertia and
correcting errors produced by the anthropogenic features definition: ARFIMA.
These type of models are considered as an extension of traditional ARIMA models,
letting the differencing parameter to take non-integer values. By doing so, ARFIMA
models are more appropriate for modeling time series with long memory 
\cite{granger_introduction_1980}. Through this work, the \texttt{arfima} function
from the R package \textit{forecast} is used \cite{hyndman_2008}. 


\subsubsection{QR probability regression}


\begin{figure*}[tbp]
\centering
\includegraphics[width=0.95\textwidth]{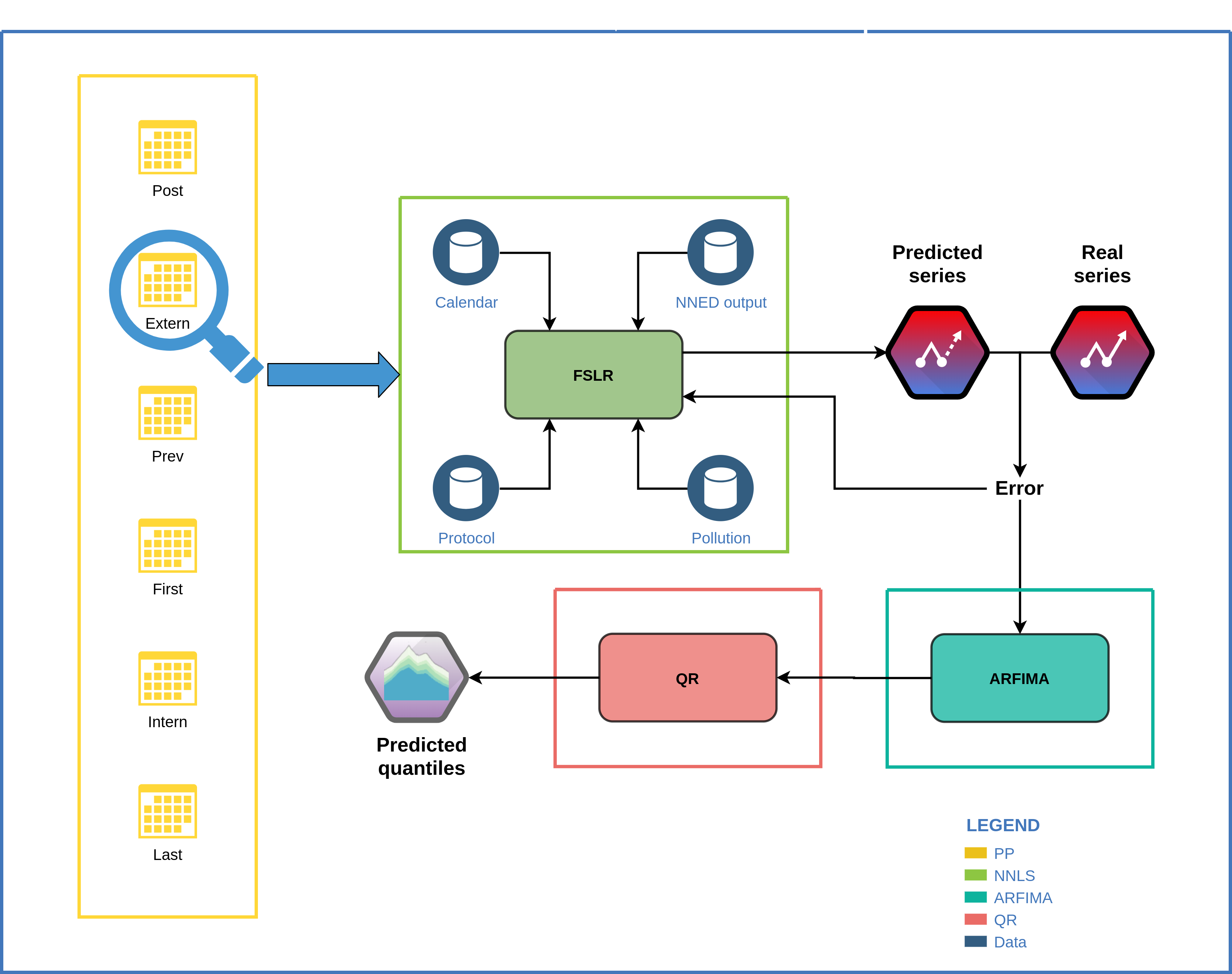}
\caption{PP-FSLR-ARFIMA-QR scheme. First, input data is classified depending
  on the type of day to be predicted (yellow). For each new series, the FSLR
  (green) takes as inputs different data sources (blue). Through linear
  relations, FSLR models different aspect of the problem, and uses its own
  prediction error for autoregulation. This same error is fed to the ARFIMA
  model (sky blue). Finally, the predicted quantiles are computed by the QR
  (red). 
  }
\label{fig:pp_nnsl}
\end{figure*}

\begin{figure*}[tbp]
\centering
\includegraphics[width=0.65\textwidth]{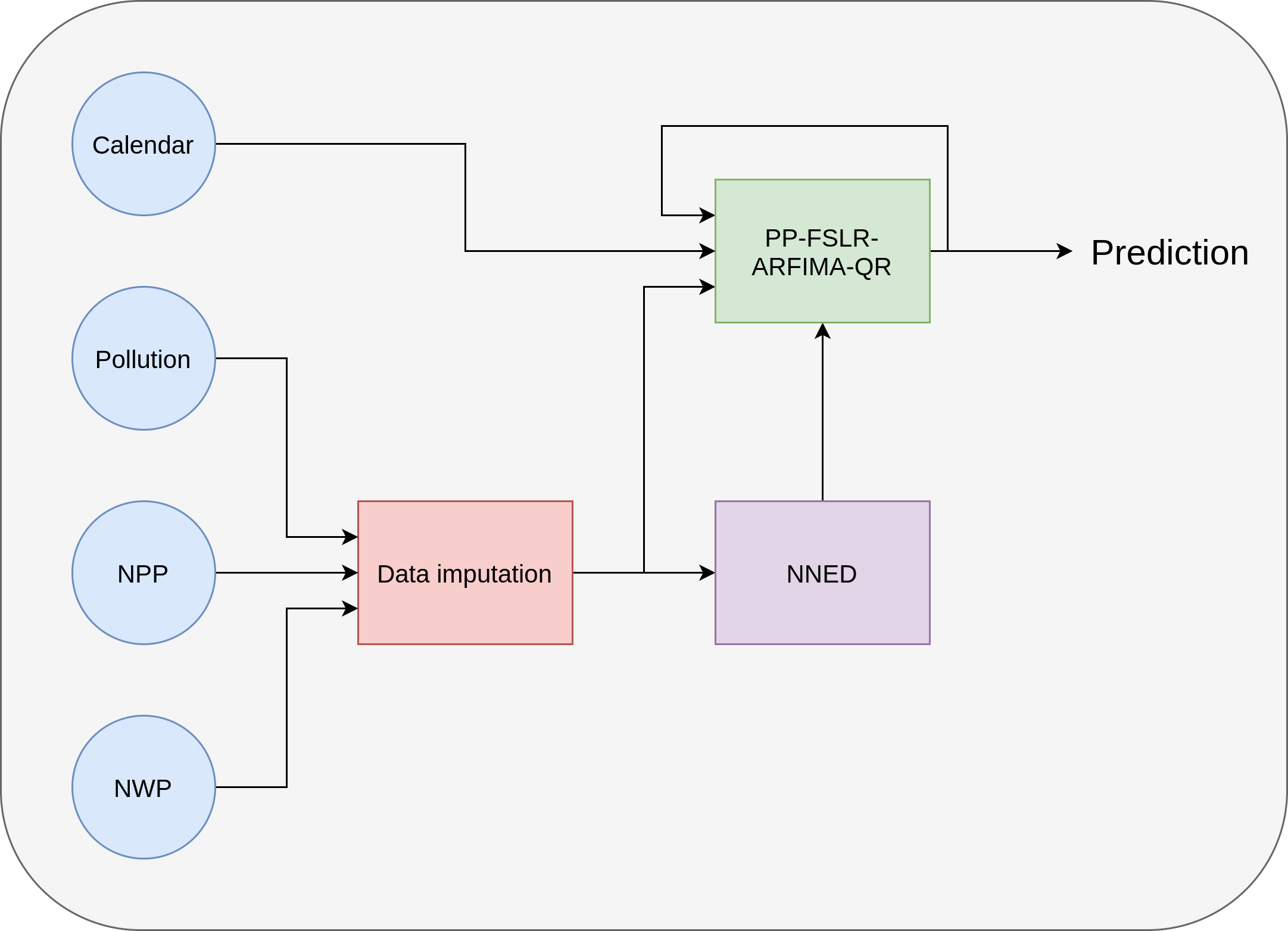}
\caption{Complete schematic of the ensemble of cascade nested models in SOCAIRE. In blue
  circles, data sources, and in squares, the mathematical components. The
  relationships between the different elements of the diagram are reflected by
  arrows. 
  }
\label{fig:com_mod}
\end{figure*}

At this point, the forecasts generated represent the mathematical expectation of
the output magnitudes. With this, one can aspire at most to asymptotically
estimate a log-normal distribution under the laws of regression. But since the
distribution will not always fit perfectly with a log-normal, it is preferable
to use a method based solely on the data.

To do this, a new probabilistic quantile regression (QR) is estimated in order
to estimate the future concentrations, with as single input the forecast of the
previous FSLR+ARFIMA model, in original terms (without applying the logarithmic
transformation), in order to obtain all percentiles from 1\% to 99\%.

In this setting, since there is not always enough contrast surface (the
data-variables ratio is low), it may happen that the estimated percentiles do
not comply with the basic rules of non-negative and non-decreasing applicable to
every probability distribution. Usually, it is in the extremes where there are
more problems
. To alleviate this inconsistency, an I-spline interpolation is
applied to these percentiles to ensure that these properties are as close as
possible to the estimated values.

A general schematic of the PP-FSLR-ARFIMA-QR model is presented in Fig.
\ref{fig:pp_nnsl}, while Fig. \ref{fig:com_mod} summarises the complete
model with the data sources that govern the system.

\section{Probabilistic prediction of the alert levels}
\label{S5}

As described in Section \ref{S2.2}, the activation of the \no protocol depends
on meeting a number of requirements, defined in three alert levels. From a
probabilistic point of view, these requirements can be seen as compound events,
and being able to compute the future probability for the activation of each
level is of utmost importance for decision makers.

According to the \no protocol, the activation of the different levels depends on
what happens in several stations at the same time and in a certain number of
consecutive hours. In order to compute the aggregated probability, the
evaluation of the probability of the intersection of several events is thus
needed, knowing only the marginal percentiles and the historical residues left
by each of the models.

\subsection{Empirical marginal distribution of the different stations}
\label{S5.1}



As we have shown above, the model for each station offers a probabilistic
forecast condensed in a quantile vector. Specifically, the 99 integer
percentiles are taken, that is, those corresponding to the probabilities
$p_{k}=1\%, 2\%, \dots , 98\%, 99\%$.

In this section, we will look for a way to calculate the marginal distribution
function for the forecast of each pollutant concentration from these quantiles
calculated by each station's model. For this, it will be necessary to calculate
the inverse of this distribution function and some statistics such as the mode,
which in turn requires an analytical representation that allows us to obtain its
first and second derivatives. In summary, we need a pair of easily computable,
continuous, and doubly derivable functions that allow us to evaluate very
efficiently and precisely approximations of the distribution function and its
inverse at any point of their respective domains. The selected method is in fact
an empirical change of variable that transforms the concentration into a
standardized normal.

We will first take into account the fact that, by definition, the estimated
quantiles are evaluations of the change in a variable that transforms the
forecasts into a uniform distribution. Although this is valid for any source
distribution, for reasons of numerical stability it is preferable to apply the
process to the logarithms of the quantiles. Thus, if we apply the inverse of the
standard normal distribution function to these log-quantiles, then the values
obtained will follow that distribution by construction. Note that the
calculation of the mode and deviation becomes trivial in this context.

During the approximation process, we will establish the restriction that the
probability density of the concentration forecast is always unimodal, which
agrees perfectly with the analyzed observations and the type of models used.

Let us think of the moment in which decision making takes place, $t_0$, and let
us call $y_{s,t}>0$ the real concentration not yet observed in station $s$ at
future instant $t>t_0$. The model of the $s$ station will give us the
$q_{s,t,k}$ percentiles of the forecast such that $P\left[y_{s,t}\leq
  q_{s,t,k}\right]=p_{k}$. The transformed values are thus defined as
$z_{s,t}=\log\left(y_{s,t}\right)$ and the standardized normal quantile is
$u_{k}=\Phi^{-1}_{0,1}\left(p_{k}\right)$, where obviously $\Phi_{0,1}$ is the
normal distribution function with mean $0$ and deviation $1$.

Now we will interpolate the pairs $\left(u_{k},z_{t,s}\right)$ by means of a
function $f_{s,t}:\mathbb{R}\longrightarrow\mathbb{R}$ that passes through those
points
\begin{equation}
f_{s,t}\left(u_{k}\right)=z_{t,s}
\end{equation}
and, in an analogous way, the inverse function
$g_{s,t}=f_{s,t}^{-1}:\mathbb{R}\longrightarrow\mathbb{R}$ will be constructed
as the interpolating function that passes through the points
$\left(z_{t,s},u_{k}\right)$. That is to say
$g_{s,t}\left(z_{k}\right)=u_{t,s}$.

This allows us to construct an approximation of the concentration distribution
function as follows:
\begin{equation}
  \Phi_{0,1}\left(g_{s,t}\left(\log\left(y\right)\right)\right)\simeq P\left[y_{s,t}\leq y\right]=\Psi_{s,t}\left(y\right).
\end{equation}
And similarly we will obtain the approximation of its inverse:
\begin{equation}
\exp\left(f_{s,t}\left(\Phi_{0,1}^{-1}\left(p\right)\right)\right)\simeq\Psi_{s,t}^{-1}\left(p\right).
\end{equation}

Although we could have directly interpolated these functions, which are the true
objective, numerically speaking the interpolation with these transformations is
more stable (largely because both $z_{s,t}$ and $u_{k}$ are not bounded).

To avoid problems in the tails of the distribution, and taking into account that
both functions are monotonous, it is highly recommended to use an interpolation
method that guarantees this monotonicity. In particular, a monotonic spline
interpolation has been used in this work. The monotony of the functions
$f_{s,t}$ and $g_{s,t}$, together with the monotony of the logarithm and the
exponential functions, guarantees that the maximum probable value of the
concentration will be $\hat{y}_{s,t}=\exp\left(f_{s,t}\left(0\right)\right)$.

Let the standardized residue of the forecast be
\begin{equation}
  \varepsilon_{s,t}=g_{s,t}\left(\log\left(y_{s,t}\right)\right)\sim N\left(0,1\right),
\end{equation}
and note that indeed, if the probable maximum forecast is exact, i.e. if
$y_{s,t}=\hat{y}_{s,t}$, then
\begin{equation}
  \varepsilon_{s,t}=g_{s,t}\left(\log\left(\exp\left(f_{s,t}\left(0\right)\right)\right)\right)=g_{s,t}\left(f_{s,t}\left(0\right)\right)=0
\end{equation}
Similarly, if the standardized residue is zero, then the forecast is exact.

\subsection{Empirical joint distribution}
\label{S5.2}

\begin{figure}[tbp]
 \centering
 \includegraphics[width=.90\textwidth]{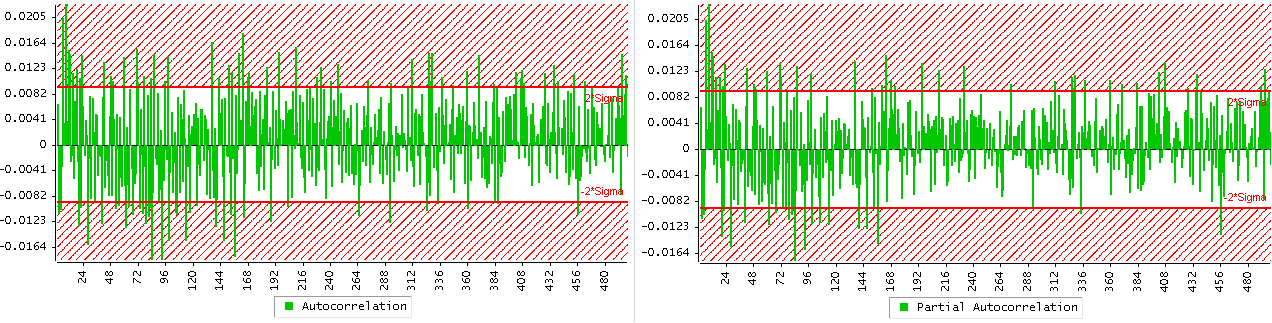}
 \caption{ACF and PACF of residuals in station 58.}
 \label{fig:stdNormResAcf}
\end{figure}


\begin{figure}[tbp]
 \centering
 \includegraphics[width=.78\textwidth]{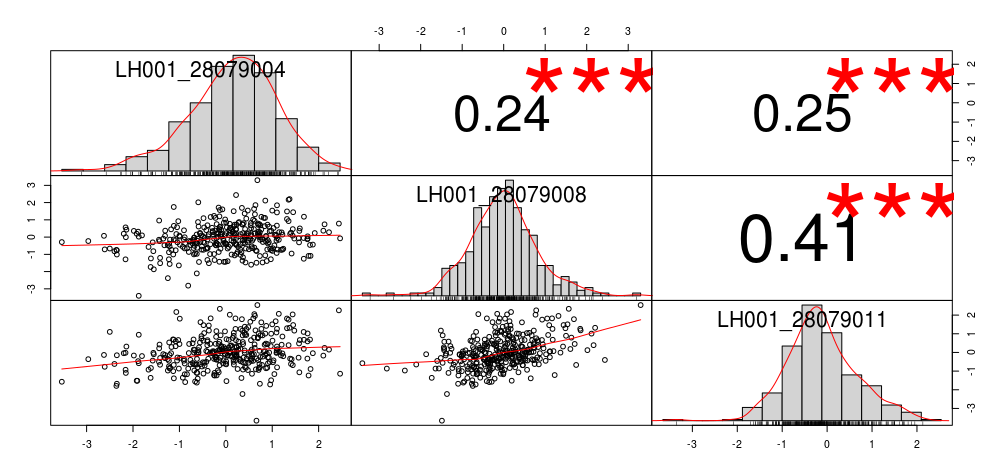}
 \caption{Instantaneous correlations and residual histograms for stations 4, 8
   and 11.}
 \label{fig:stdNormResCor}
\end{figure}

\begin{figure}[tbp]
 \centering
 \includegraphics[width=.7\textwidth]{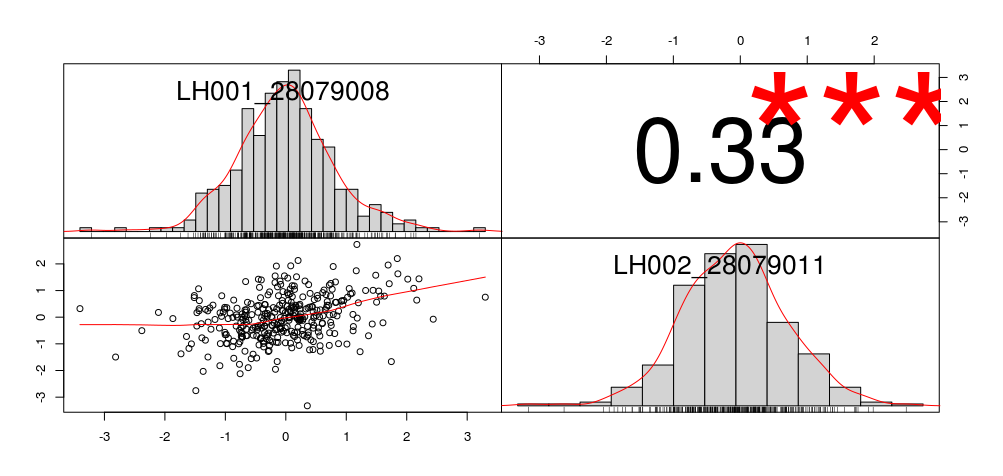}
 \caption{Summary of the 1 hour ahead cross-correlations of the residuals series
   for station 8 and 11.}
 \label{fig:stdNormResCorH1}
\end{figure}


Section \ref{S4} has described the models that marginally predict the
concentration of each pollutant at each station for different time horizons.
These models, thanks to their ARIMA structure, are able to adequately treat the
internal temporal correlation of each station, that is, the autocorrelation of
each of the series of pollutants of the different stations. In Fig.
\ref{fig:stdNormResAcf}, it can be seen that the autocorrelation function (ACF)
is never too big, and that when it does exceed the 2 sigma limits, so does the
partial autocorrelation function (PACF). This fact suggests that these are
spurious correlations or any other types of concurrent causes, not linked to
time.

However, in view of Fig. 
\ref{fig:stdNormResCor},
there is nothing that indicates that residuals from different stations will be
independent of each other. Rather, they appear to correlate.

On the one hand, even if NNED models the spatio-temporal dynamics of the
process, it is expected that closer stations will be more similar amongst them,
giving rise to positive correlations between their residues. On the other hand,
as shown in Fig. \ref{fig:stdNormResCorH1}, the errors in each forecast horizon
for a single station will also not be independent of other stations' previous
horizons. In fact, this occurs mostly mutually, present errors of a station
correlate with the past errors of another station and vice versa.

In the previous section we have seen how to obtain, by means of an interpolative
variable change, standardized normal residues in a marginal way for each station
$s$ and for each future instant $t$ at current time $t_0$. However, if the
independence hypothesis is not plausible it is clear that knowing the marginal
distributions does not imply knowing the joint distribution. 

A family of models which are naturally capable of dealing with this situation
are the X-VecARMA
, a type of multivariate models \cite{sims1980macroeconomics} that
include exogenous inputs, cointegration, and vector ARMA. They are considered
very powerful for the representation of cross-correlated vector processes that
might include exogenous factors eventually shared by several of them. However,
they are intractable in computational terms for this setting.

Thus, we propose an empirical multi-normal copula \cite{nelsen_introduction_1999}
to approximate the joint distribution for every station and
horizon. The aim is to obtain an estimate of such joint distribution function
for all the forecasts obtained marginally, both in time and space, using the
joint sample correlation matrix between each pair of stations among all the
horizons and stations.

However, since there are $48$ horizons and $24$ stations, that gives us a square
matrix of $1152$ rows, and we would need at least $10$ years of forecasts to
obtain a meager $3$ to $1$ response surface, which is clearly unacceptable. For
this reason, we have developed a boxed tridiagonal scheme, in which correlations
are only taken into account one period ahead. With this scheme, only one year of
forecasting is sufficient to obtain a reasonable estimate.

We will assume that the joint distributions of these standardized residues only
depend on the station and the forecast horizons $h=t-t_0$ and $h-1=t-1-t_0$, but
not on the specific moment $t$, since the forecasts will be made every day at
the same time.

Since the marginal distributions of all the $\varepsilon_{1,t}$ are normal,
unbiased and with unit variance, the joint distribution of all stations,
\begin{equation}
  \varepsilon_{h}=\left(\varepsilon_{1,t_{0}+h},\cdots,\varepsilon_{s,t_{0}+h},\cdots,\varepsilon_{S,t_{0}+h}\right)^{T}\in\mathbb{R}^{S},
\end{equation}
will be an unbiased multinormal with an unknown but obligatory unitary
covariance matrix, that is, equal to the correlation matrix. In the same way we
will suppose that the residuals $\left(\varepsilon_{h-1},\varepsilon_{h}\right)$
corresponding to each pair of consecutive horizons are also distributed the same
way.

By the principle of causality, for the previous horizon, $\varepsilon_{h-1}$, an
independent distribution of the following $\varepsilon_{h-1}$ will be
postulated, since future events cannot influence the past. In this way, we can
define the joint distribution of the different stations in each horizon in a
recursive way:
\begin{equation}
  \begin{array}{c}
    \varepsilon_{1}\sim \mathcal{N}(0,C_{1})\\
    \left(\varepsilon_{h-1},\varepsilon_{h}\right)\sim \mathcal{N}(0,C_{h}), \forall h=2,3,\ldots,H\\
    C_{h}=\left(\begin{array}{ll}
                  C_{h-1,h-1} & C_{h-1,h}\\
                  C_{h-1,h}^{T} & C_{h,h}
                \end{array}\right)\in\mathbb{R}^{2S\times2S}, \forall h=2,3,\ldots,H\\
    C_{1}=C_{1,1}, C_{h-1,h}, C_{h,h}\in\mathbb{R}^{S\times S}\\
    C_{h-1,h-1,s,s}=C_{h,h,s,s}=1\\
    C_{h,h,s,s'}=\rho_{\varepsilon_{s,t_{0}+h},\varepsilon_{s',t_{0}+h}}\in\left(-1,1\right)\\
    C_{h-1,h,s,s'}=\rho_{\varepsilon_{s,t_{0}+h-1},\varepsilon_{s',t_{0}+h}}\in\left(-1,1\right)
  \end{array}
\end{equation}

Note that the joint distribution of all horizons would have a tridiagonal
covariance matrix with partitions of order $S$:
\begin{equation}
\mathrm{C}=\left(\begin{array}{cccc}
C_{1,1} & C_{1,2} &  & 0\\
C_{1,2}^{T} & \ddots & \ddots\\
 & \ddots & \ddots & C_{H-1,H}\\
0 &  & C_{H-1,H}^{T} & C_{H,H}
\end{array}\right).
\end{equation}

If we calculate the forecasts for enough dates $t_0$ of the past, at the same
time of the day and with the same horizons $h=1,2,\dots,H$, we can obtain many
samples of the residues with which we can thus estimate the matrices $C_{h-1,h}$
and $C_{h,h}$. In this way, we would obtain the distributions for each horizon
conditioned on the previous horizon, using the formula known analytically for
the conditional partitioned multivariate normal:
\begin{equation}
  \left.
    \begin{array}{c}
      \varepsilon_{h}\sim \mathcal{N}\left(\mu_{h},C'_{h}\right)\\
      \mu_{h}=C_{h-1,h}^{T}C_{h-1,h-1}^{-1}\varepsilon_{h-1}\in\mathbb{R}^{S}\\
      C'_{h}=C_{h,h}-C_{h-1,h}^{T}C_{h-1,h-1}^{-1}C_{h-1,h}\in\mathbb{R}^{S\times S}
    \end{array}\right\} \forall h=2,3,\ldots,H
\end{equation}

These matrices can be stored for later use in future joint forecasts, along with
their Cholesky and inverse decompositions:
\begin{equation}
  \begin{array}{c}
    C_{h,h}=L_{h}L_{h}^{T}, \; \forall h=1,2,3,\ldots,H\\
    C'_{h}=L'_{h}L'^{T}_{h}, \; \forall h=2,3,\ldots,H
  \end{array}
\end{equation}

First we simulate $N$ vectors of $N$ standardized independent residuals for the
first horizon
\begin{equation}
  \eta_{1,n}\sim \mathcal{N}\left(0,I\right), \forall n=1,2,\ldots,N
\end{equation}
and pre-multiplying them by $L_{1}$ we will have the standardized residuals of
all the stations for the first horizon:
\begin{equation}
  \varepsilon_{1,n}=L_{1}\eta_{1,n}\sim \mathcal{N}\left(0,C_{1}\right).
\end{equation}

From there, also starting from independent residuals
\begin{equation}
  \eta_{h,n}\sim \mathcal{N}\left(0,I\right), \forall n=1,2,\ldots,N,
\end{equation}
residuals of each horizon conditioned by the previous one can be simulated:
\begin{equation}
  \varepsilon_{h}=\mu_{h}+L_{h}'\eta_{h,n}.
\end{equation}

On the one hand, this approach solves the problem of time correlation in
consecutive hours, which is what is required, and on the other hand it is simple
enough to be able to generate correct estimations.

Finally, applying the transformations detailed Section \ref{S5.2} we obtain $N$
realizations of the future forecasts of the concentrations of the different
stations in each horizon:
\begin{equation}
  y_{s,t_{0}+h,n}=\exp\left(f_{s,t}\left(\varepsilon_{s,h,n}\right)\right).
\end{equation}

If this simulation is repeated a sufficient number of times we can calculate any
joint statistic from the forecasts of the concentrations in the different
stations. In particular, for example, to calculate the probability of activation
of the pre-warning level of the \no protocol, defined as the probability of the
concentration of \no exceeding a certain threshold $\varUpsilon=180$ in at least
two stations during two consecutive hours, it will simply be necessary to
calculate what proportion of the simulated samples meet these criteria.

\section{Operation and performance}
\label{S6}

\subsection{Operation}
\label{S6.1}

\begin{figure}[tbp]
 \centering
 \includegraphics[width=1\textwidth]{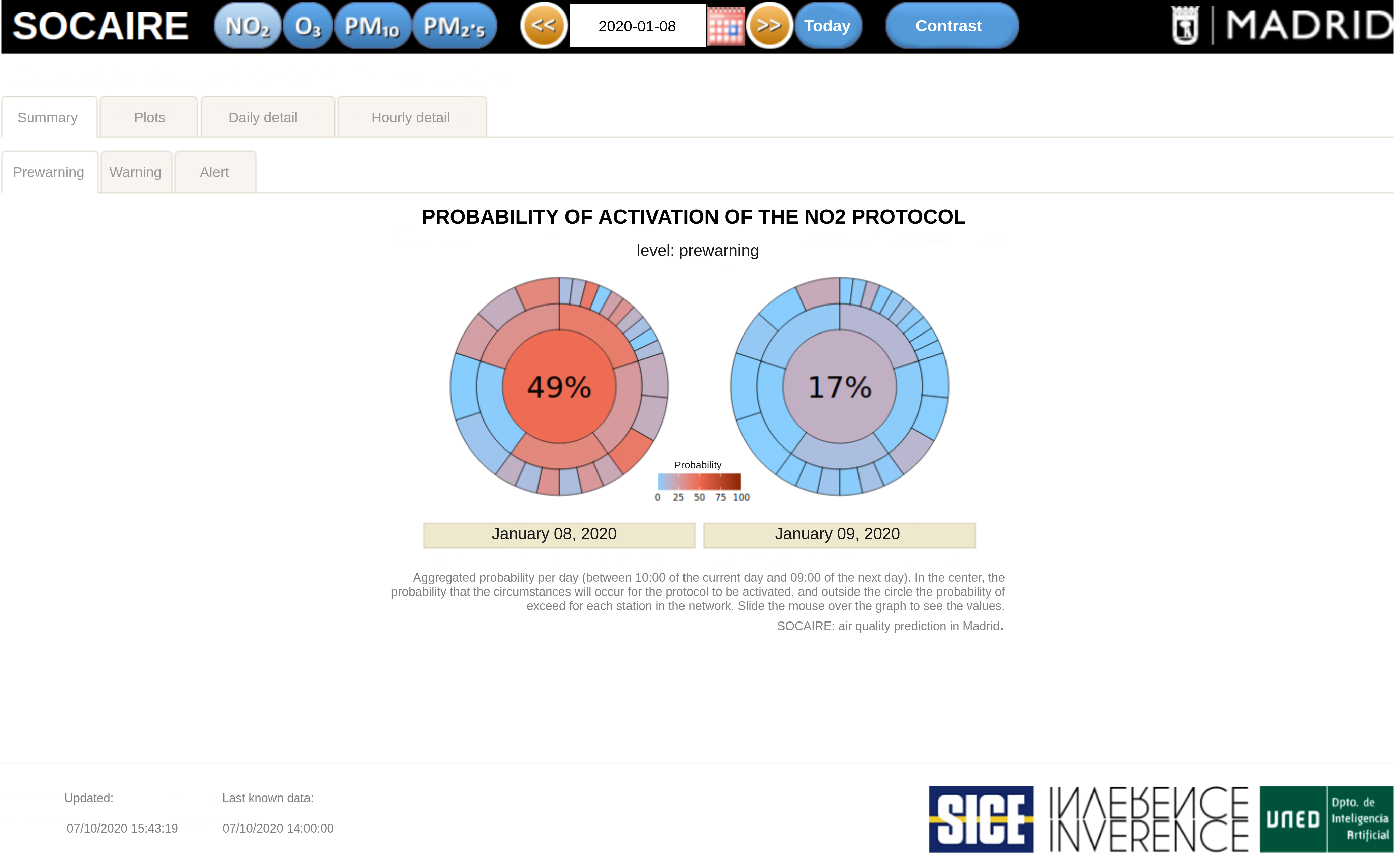}
 \caption{Main page of the SOCAIRE web app for controlling and monitoring
   pollution in the city of Madrid. 
     }
 \label{fig:app_1}
\end{figure}

\begin{figure}[tbp]
  \centering
  \begin{subfigure}{0.5\columnwidth}
    \includegraphics[width = 1\textwidth]{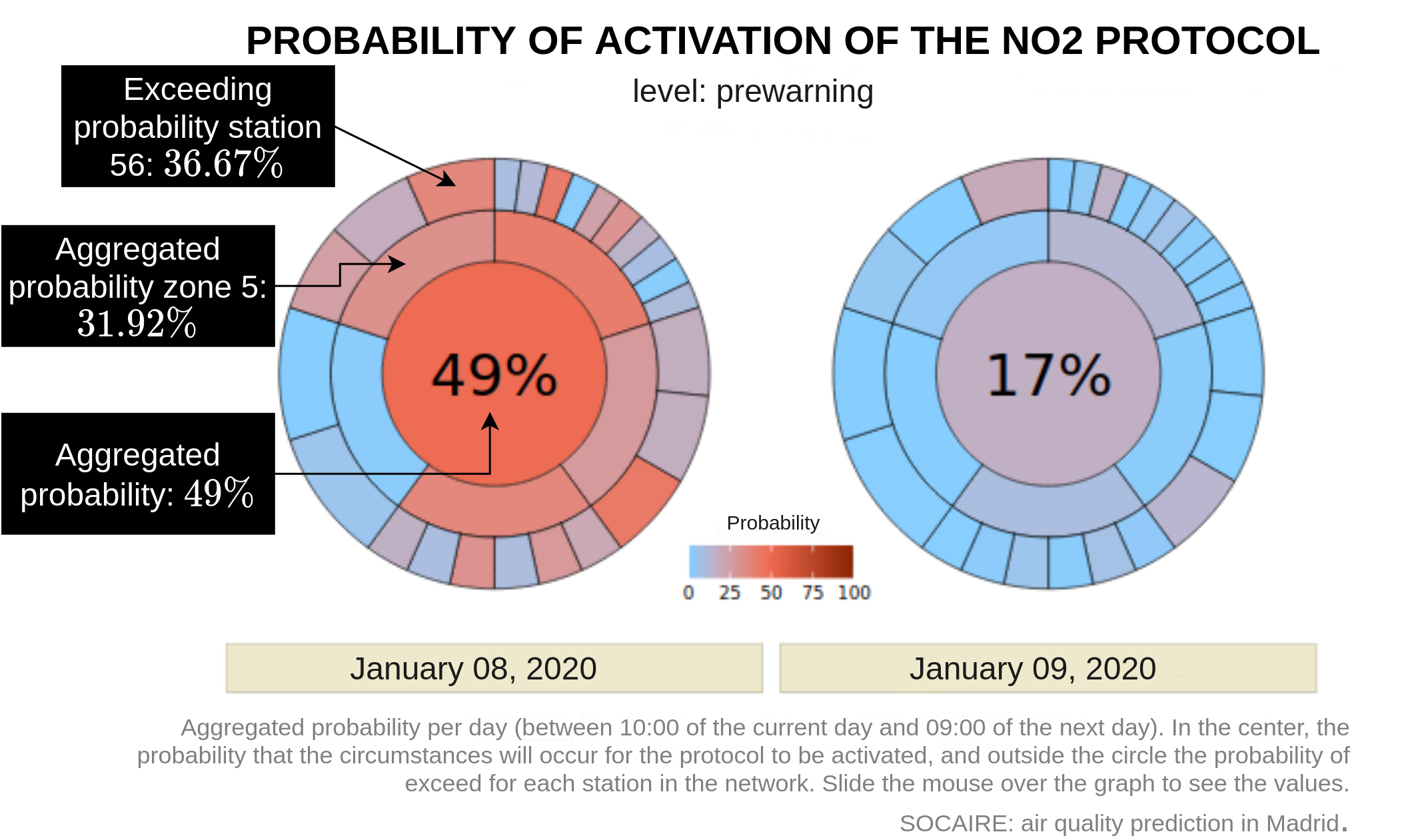}
    \caption{Prewarning.}
    \label{fig:app_2_1}
  \end{subfigure}%
    ~
  \begin{subfigure}{0.5\columnwidth}
    \includegraphics[width = 1\textwidth]{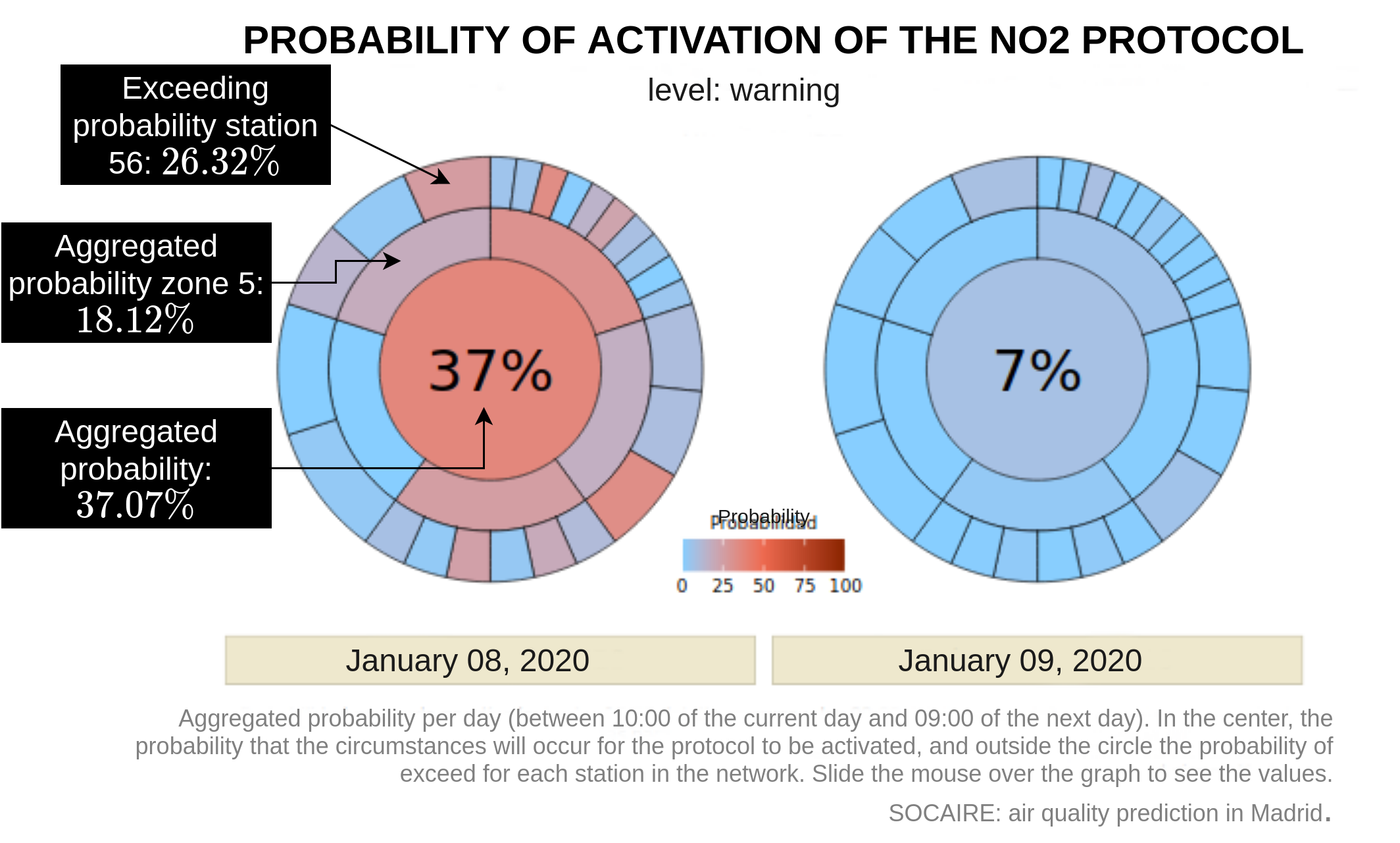}
    \caption{Warning.}
    \label{fig:app_2_2}
  \end{subfigure}
    ~
  \begin{subfigure}{0.5\columnwidth}
    \centering
    \includegraphics[width = 1\textwidth]{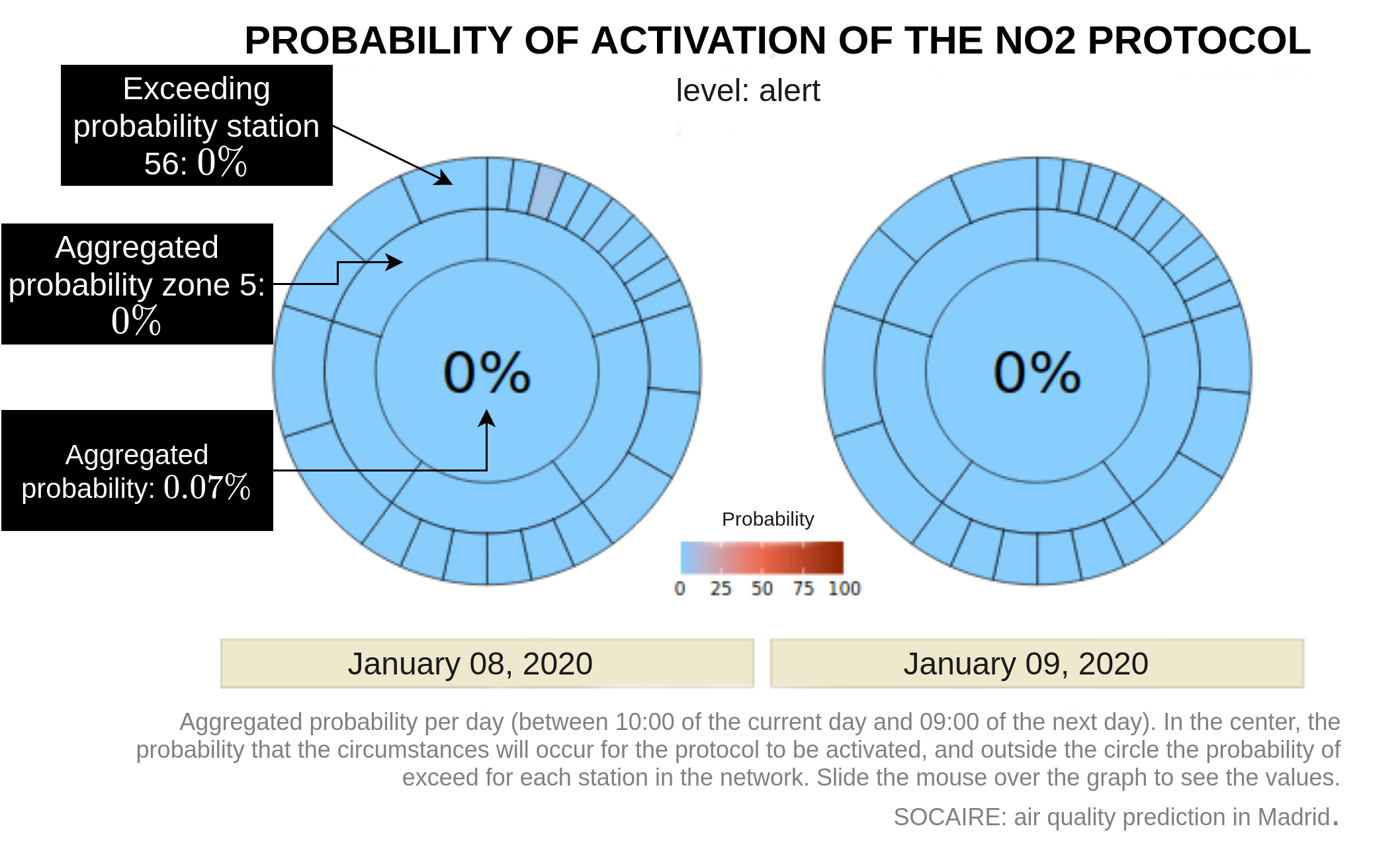}
    \caption{Alert.}
    \label{fig:app_2_3}
   \end{subfigure}%

  \caption{Probability of activation for the three levels of the \no protocol,
    predicted the 29/09/2020. 
      }
  \label{fig:app_2}
\end{figure}

\begin{figure}[tbp]
  \centering
  \begin{subfigure}{0.5\columnwidth}
    \includegraphics[width = 1\textwidth]{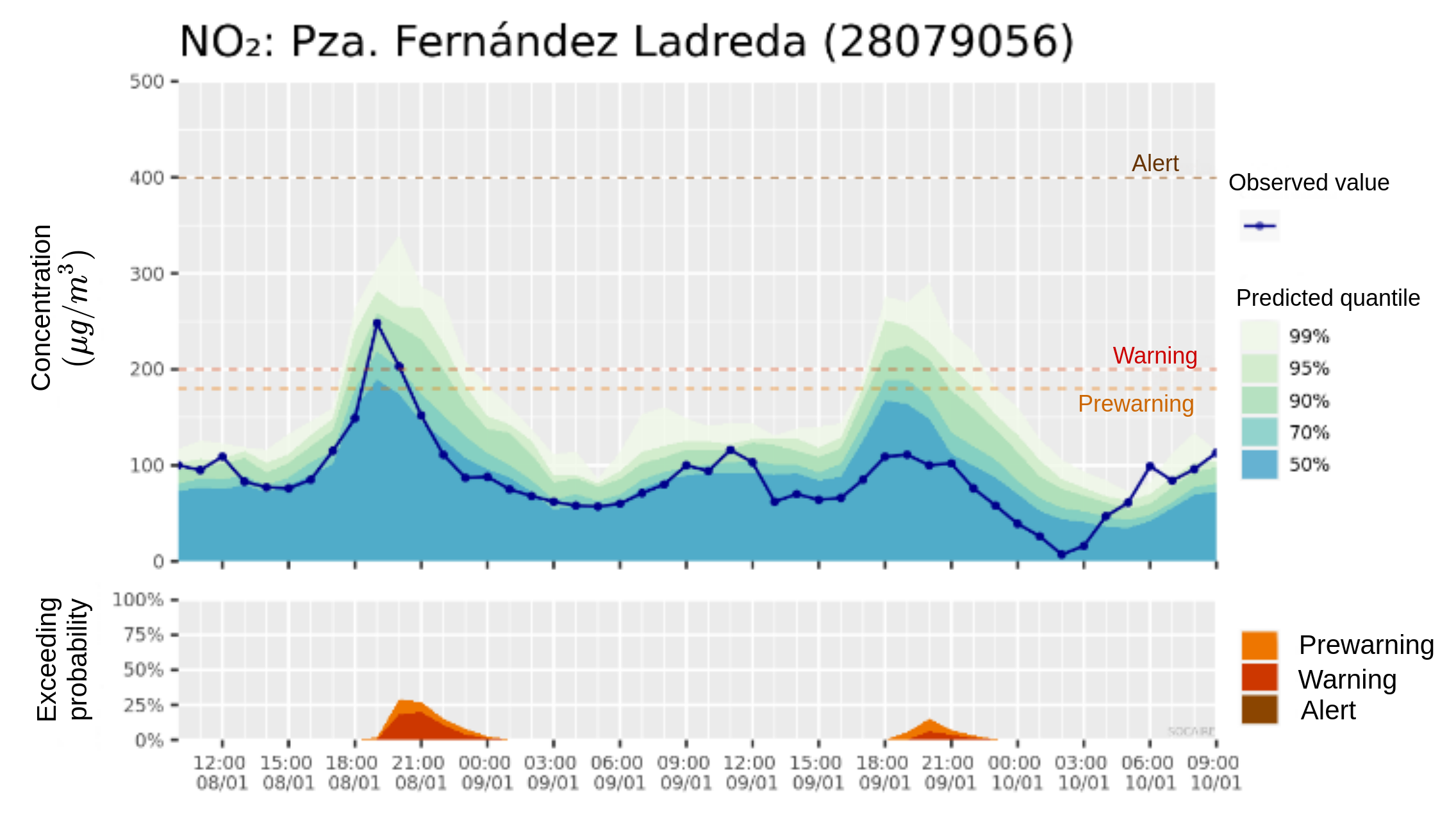}
    \caption{\no.}
    \label{fig:app_no2}
  \end{subfigure}%
    ~
  \begin{subfigure}{0.5\columnwidth}
    \includegraphics[width = 1\textwidth]{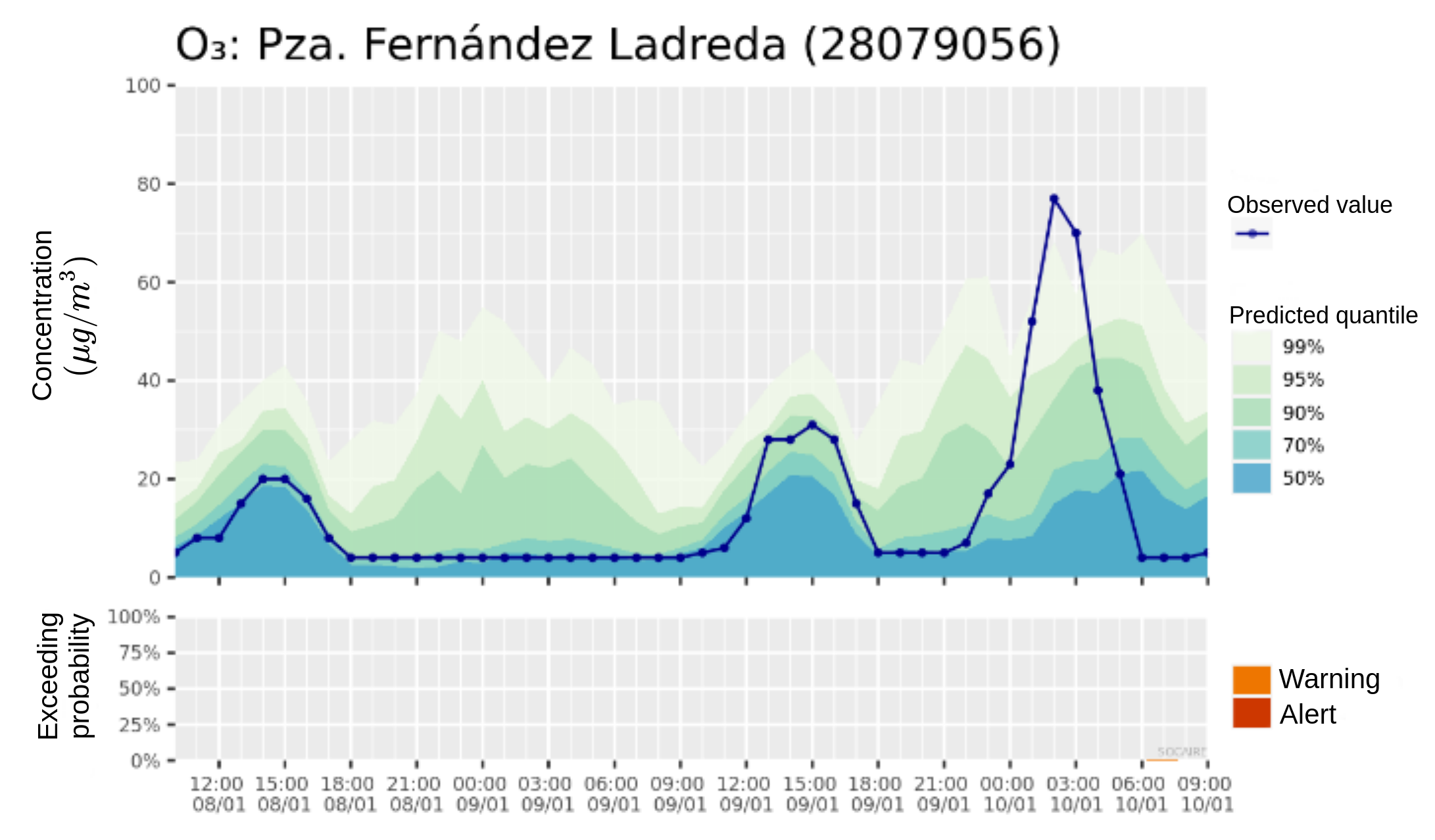}
    \caption{O\textsubscript{3}.}
    \label{fig:app_o3}
  \end{subfigure}
    ~
  \begin{subfigure}{0.5\columnwidth}
    \centering
    \includegraphics[width = 1\textwidth]{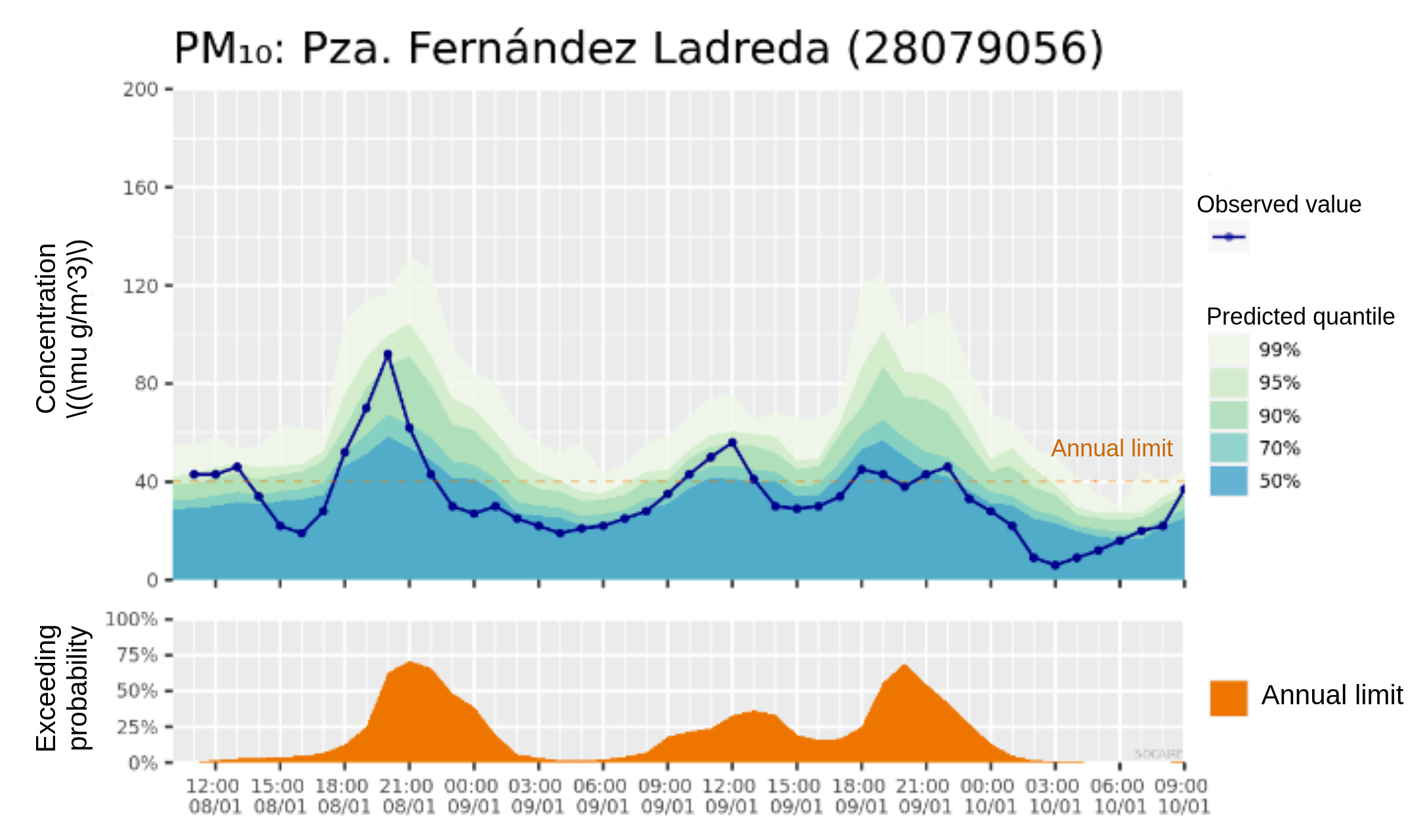}
    \caption{PM$10$.}
    \label{fig:app_pm10}
   \end{subfigure}%
    ~
  \begin{subfigure}{0.5\columnwidth}
    \centering
    \includegraphics[width = 1\textwidth]{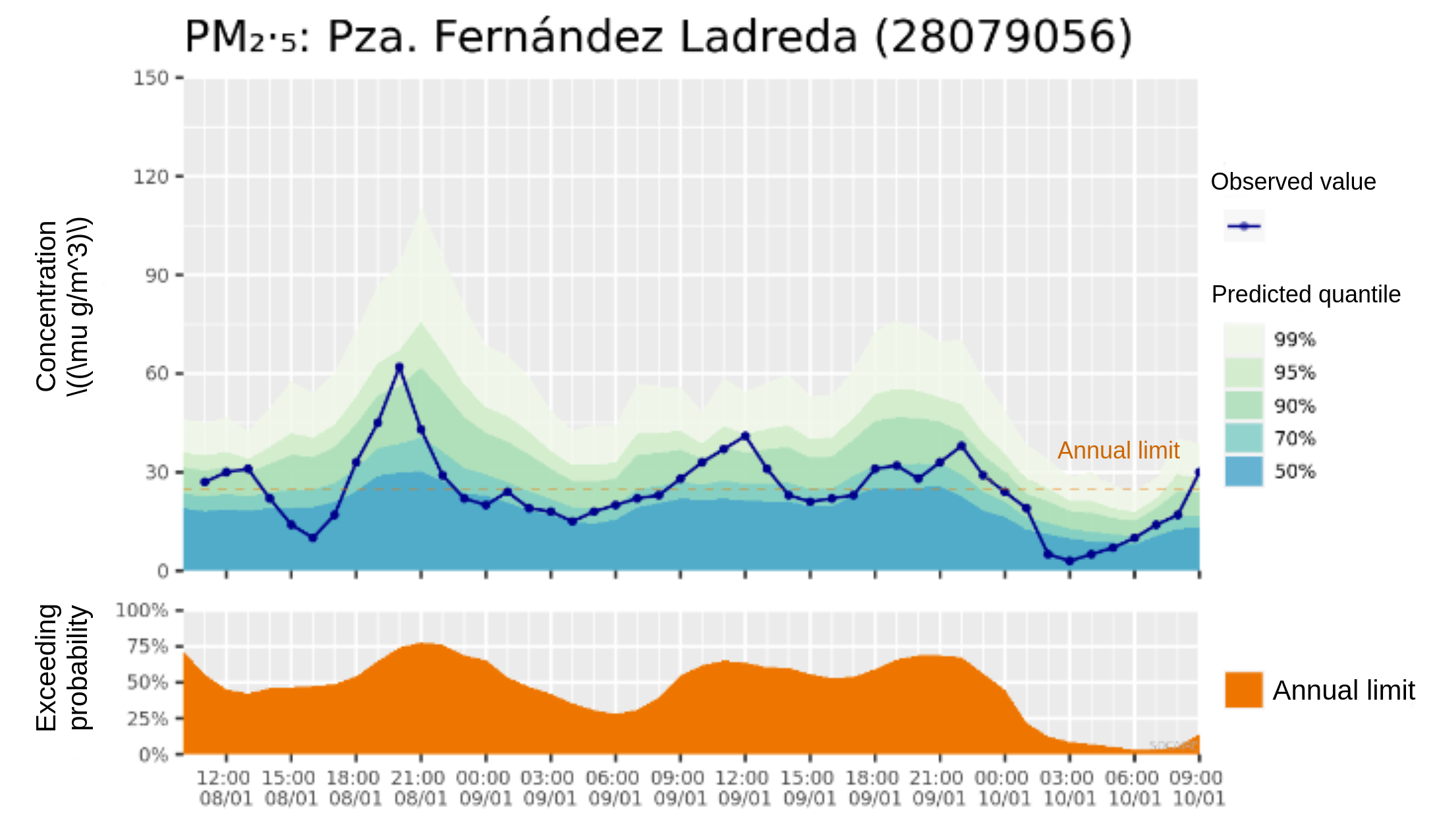}
    \caption{PM$2.5$.}
    \label{fig:app_pm25}
  \end{subfigure}
  
  \caption{Forecast quantile and probabilities of exceedance for each hour in
    station 56 for 08/01/2020. The real observed values are represented by the
    blue lines and dots and, thus, it is possible to have a reference about
    SOCAIRE performance (which will be covered in Section \ref{S6.2}).
      }
  \label{fig:app_3}
\end{figure}

In order to be used by decision makers in the department of the city council in
charge of air quality, SOCAIRE has been integrated with a web app that allows to
simply and directly view the forecasts for pollutants and the probability of
reaching the levels established within the \no protocol as explained in section
\ref{S2.2}. This section will show the site structure and its basic operation
principles.

The main overview of the web tool is shown in Fig. \ref{fig:app_1}. On the one
hand, at the top you can choose the pollutant to display (blue buttons), the
date on which you want to make a query (calendar button), and different submenus
where you can see in more detail the probability that the protocol will be
activated (shown tab), and both the system predictions and a summary of contrast
measures. On the other hand, in the central part the information related to the
submenu in which the user is at that moment is shown. In this specific case, the
probability of the levels of the \no protocol being activated.

The operation of the tool for monitoring the future probability of reaching the
different levels of the protocol are presented in Fig. \ref{fig:app_2}. After
using the ensemble of nested models described in Section \ref{S4} to forecast
\no quantiles, the outer rings show the probability of each individual station
exceeding the levels set in the \no protocol (180 $\mathrm{\mu g/m^{3}}$, 200
$\mathrm{\mu g/m^{3}}$, and 400 $\mathrm{\mu g/m^{3}}$ for prewarning, warning,
and alert respectively). Once the individual probabilities are computed, it is
possible to use the process explained through Section \ref{S5} to estimate
probabilities of compound events. Given that the protocol is defined over areas
and not for individual station levels, the intermediate ring shows the
probability of exceeding the expected pollution levels for each of the 5 areas
in which Madrid is partitioned in the \no protocol (see Fig. \ref{fig:zones}).
Lastly, the inner ring contains the aggregated probability of the different
levels of the protocol being activated in the entire city. It uses the
probabilities over the five areas to estimate this final probability.

Since the set of mobility measures defined in the \no protocol depends on
reaching extreme levels in various stations and for a pre-set number of
consecutive hours, having such an overview is especially important. However, it
is also interesting to visualize the individual forecast for each station over
time. The SOCAIRE website allows viewing the actual forecasts for each pollutant
and each station, as shown in Fig. \ref{fig:app_3}. Together with the predicted
quantiles and real observed values, these plots also show the probability of
exceeding each level and the levels themselves.

\subsection{Performance analysis}
\label{S6.2}

\begin{table*}[btp]
  \centering
  \caption{Average error for $t = 1$ to $t = 48$, calculated over all stations.
    For a more detailed view of error metrics distribution, see Fig.
    \ref{fig:results_1}.}
  \label{tab:res_total}
  \resizebox{\textwidth}{!} {
    \begin{tabular}{@{}rrrrrrrrr@{}}
      \toprule
      & \multicolumn{2}{l}{\no} & \multicolumn{2}{l}{$O_3$} & \multicolumn{2}{l}{$PM10$} & \multicolumn{2}{l}{$PM2.5$} \\ 
      \cmidrule(l){2-3}
      \cmidrule(l){4-5}
      \cmidrule(l){6-7}
      \cmidrule(l){8-9}
      
      & RMSE            & Bias              & RMSE              & Bias             & RMSE             & Bias            & RMSE            & Bias \\ \midrule
      CAMS        & $23.5 \pm 9.1$  & $12.3 \pm 9.8$    & $19.1 \pm 5.0$    & $-3.2 \pm 6.2$   & $13.9 \pm 3.7$   & $6.3 \pm 3.7$   & $6.5 \pm 1.3$   & $1.1 \pm 2.1$ \\
      Persistence & $26.4 \pm 9.3$  & $-1.4 \pm 3.5$    & $27.4 \pm 4.8$    & $0.5 \pm 16.0$   & $15.5 \pm 3.1$   & $-0.6 \pm 3.5$  & $7.8 \pm 1.4$   & $-0.3 \pm 1.4$  \\ 
      SOCAIRE     & $14.9 \pm 4.8$  & $-0.2 \pm 0.8$    & $15.8 \pm 2.8$    & $1.6 \pm 1.0$    & $10.6 \pm 2.5$   & $0.3 \pm 0.7$   & $5.4 \pm 1.0$   & $-0.1 \pm 0.65$ \\ \bottomrule
    \end{tabular}
  }
\end{table*}

\begin{figure}[tbp]
 \centering
 \includegraphics[width=1\textwidth]{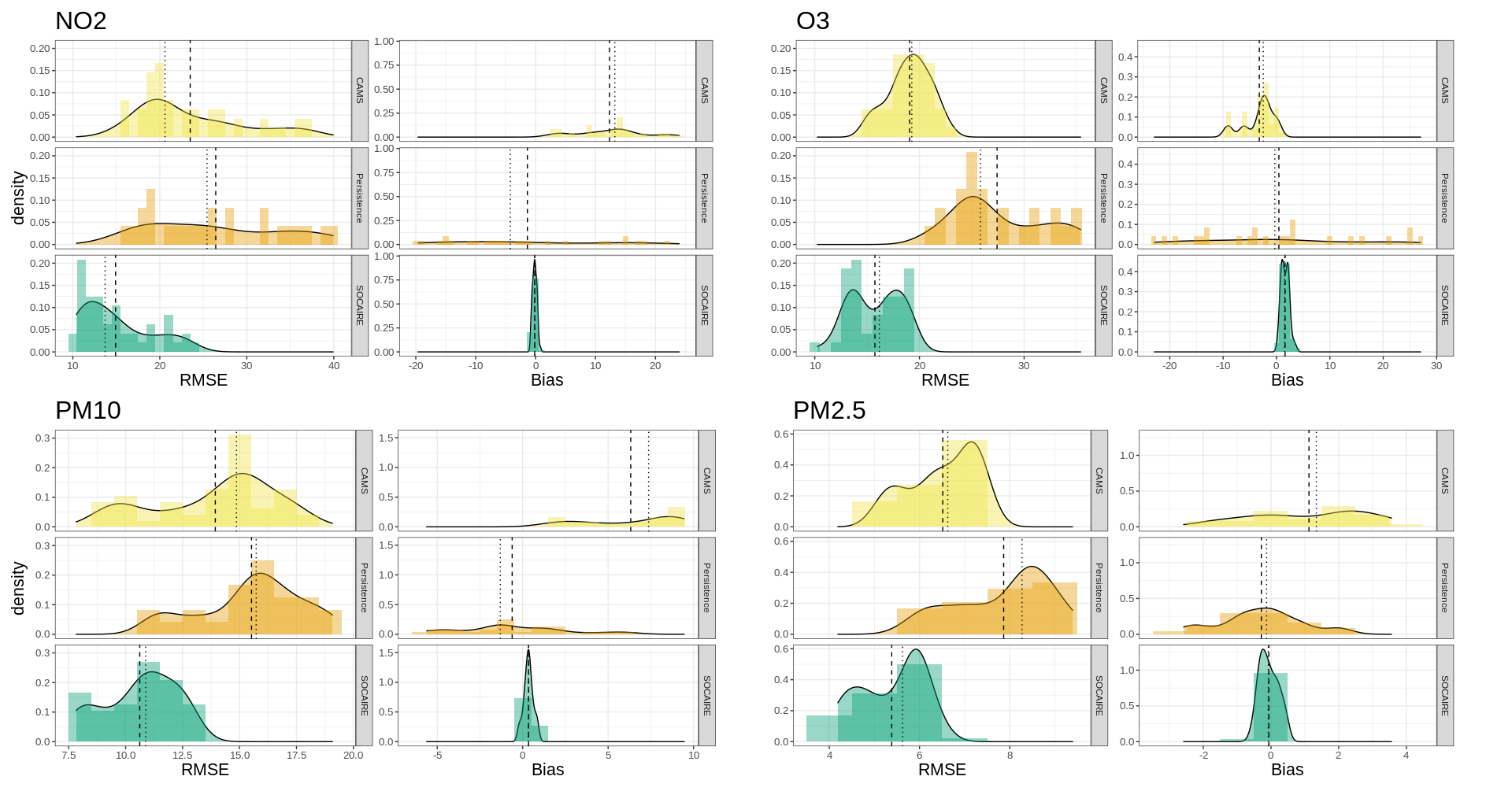}
 \caption{Error distribution for the four pollutants in terms of RMSE and bias.
   Dashed vertical line represents the mean, dotted vertical line represents the
   median.}
 \label{fig:results_1}
\end{figure}

\begin{figure}[tbp]
 \centering
 \includegraphics[width=1\textwidth]{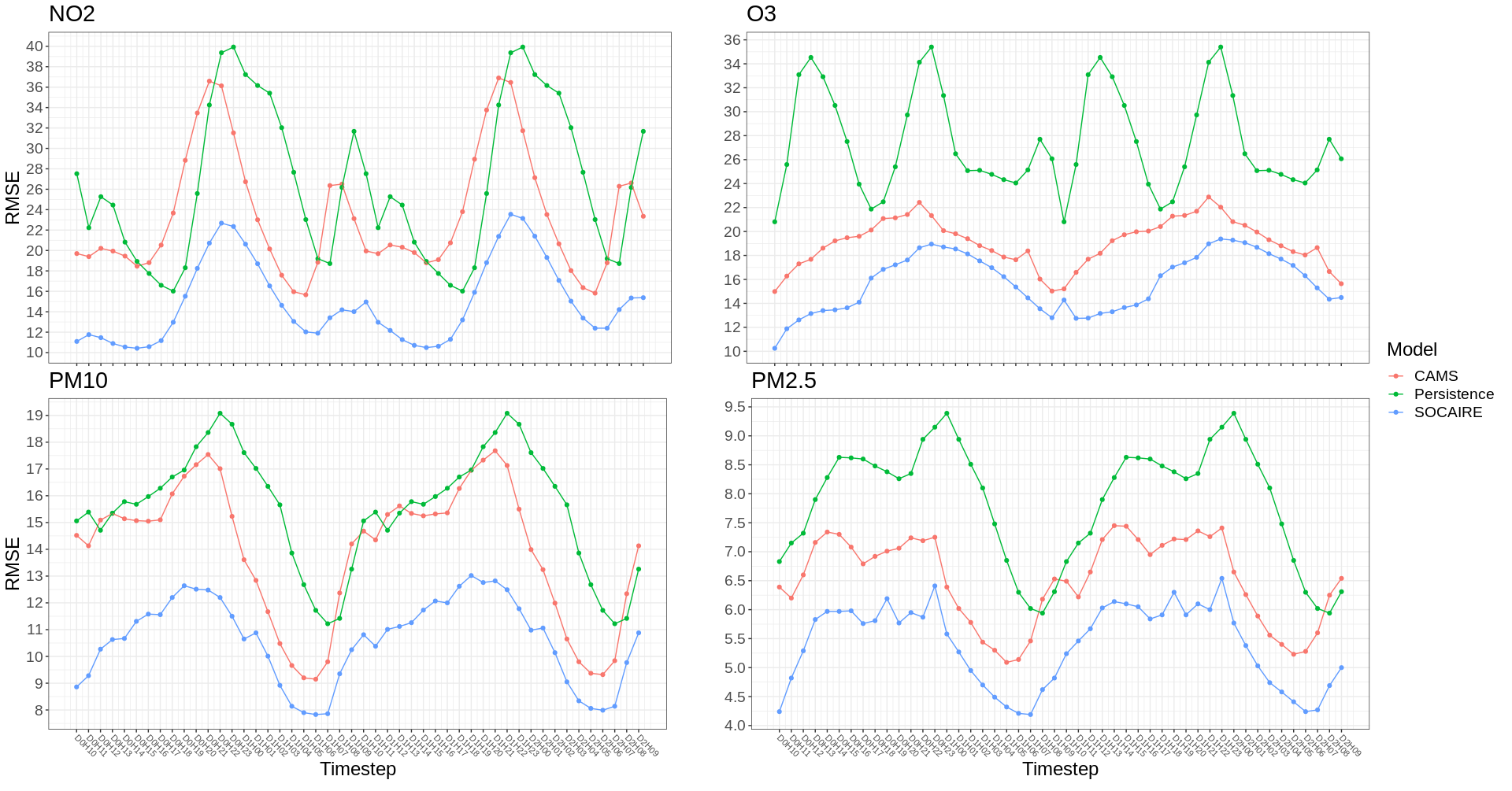}
 \caption{RMSE error by timestep. D makes reference to the day and H to the hour
   (D0H10 means day 0 or present day at 10:00).}
 \label{fig:results_2}
\end{figure}

\begin{figure}[tbp]
 \centering
 \includegraphics[width=1\textwidth]{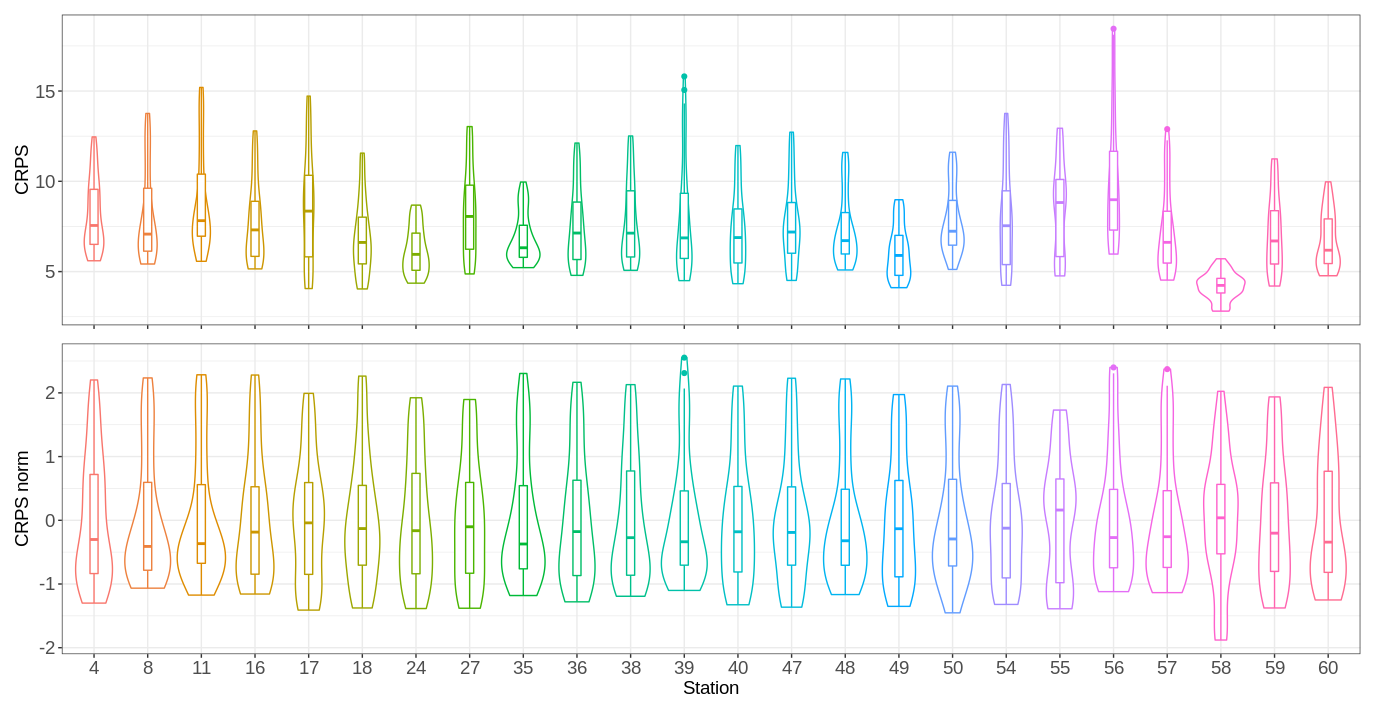}
 \caption{Comparison of the error distribution for all stations (top). Scaling CRPS
   to a $\mathcal{N}(0,1)$ let us conclude that all stations have been correctly modeled
   in the spatial dimension (bottom).
}
 \label{fig:results_3}
\end{figure}


Usual error metrics, as RMSE, refer to expected values, which are found in the
central part of the distribution, but do not take into account any other
information, and are thus particularly unfit to evaluate probabilistic
forecasts. Since the most usual models produce point forecasts and not the
entire distribution, these kinds of metrics are the only option. However, when
dealing with the prediction of the complete distribution as in our case, other
metrics have been proposed in order to summarise model performance information
in a more comprehensive and realistic way. For example, CRPS is a measure of the
squared difference between the forecast cumulative distribution function (CDF)
and the empirical CDF of the observation \cite{gneiting_probabilistic_2014}.

As we will show, in terms of performance SOCAIRE compares favorably to
benchmarks. In order to get a clear and quick idea about the behavior of the
model, Table \ref{tab:res_total} shows the RMSE and bias (averaged both in time
and space) of the proposed methodology and compares it with two other models
that, due to their characteristics, make it easier to understand the real
performance of SOCAIRE: persistence and the NPP provided by CAMS.

The persistence model is a naive model in which the forecast value is taken to
be the observed value at the previous timestep. It is, thus, a good benchmark
model and one can get a rough idea of how good a new model is by seeing how much
improvement there is with respect to persistence. In our specific case, for
contractual reasons, we use a more elaborated version of persistence which
includes the daily, weekly, and annual cyclical structure of the series, and is
thus a simple although powerful model.

Similarly, the NPP provided by CAMS represent another good baseline to be
improved upon by any new model. Since it is based on a synoptic scale, it is
expected that any model focused on a smaller and concrete terrain extension will
improve its results. If this is not the case, it would make more sense to use
CAMS NPP as an approximation instead of the proposed new methodology.

For a more detailed view of error metrics, refer to Fig. \ref{fig:results_1}. As
it can be seen, SOCAIRE consistently outperforms both baselines in terms of RMSE
and bias for the four pollutants. Concretely, SOCAIRE supposes an average RMSE
improving of $32\%$ with respect to CAMS and $60\%$ with respect to persistence,
reinforcing the idea that SOCAIRE shows good performance and behaves very well
as a predictor. Also, SOCAIRE demonstrates to be in general terms an unbiased
predictor of pollution, which emphasizes the fact that the proposed model is
being able to correctly describe the aforementioned terms related to the system.

Another issue that is of special importance in our problem is the behavior of
the model depending on the prediction timestep/hour. As it was shown in Fig.
\ref{fig:time_dep}, the series are highly hour-dependent. For example, \no
presents peaks usually around 08:00--10:00 and 22:00--00:00. In the framework of air
quality management and monitoring, these peaks are extremely important as they
represent the higher risk and, consequently, the moments when the maximum
recommended and/or permitted levels are usually exceeded. Thus, and given that
one of the main objectives of SOCAIRE framework is forecasting the probability
of each level of the \no protocol, showing a good performance in peak hours is
of crucial importance.

Fig. \ref{fig:results_2} presents the RMSE error for each pollutant and for each
prediction horizon averaged over all stations. From this figure, it becomes clear
that SOCAIRE is especially efficient in peak hours, where the gap with baseline
models is even wider.

Until now, we have covered aggregated error over all stations. As the activation
of the \no protocol depends on compound events of individual stations, it is
important to make sure all of them behave similarly. As it was explained before,
the complete model has a module which is able to relate and exploit shared
spatial information (Section \ref{S4.2}), but it also models each station
independently based on its own characteristics (Section \ref{S4.3}). By taking
into account both types of information, we expect to avoid possible biases of
predominance by some spatial areas over others but still be able to make use of
the relations that exist among them. The CRPS for the \no predictions at each
station is shown in the top row of Fig. \ref{fig:results_3}. It is worth noting
that stations with lower CRPS errors correspond to green areas of the city of
Madrid (Stations 24, 49, and 58). Scaling these CRPS values to a
$\mathcal{N}(0,1)$ (bottom row of Fig. \ref{fig:results_3}) let us see how all
error distributions have a very similar behavior. Hence, it is possible to
assure that our modeling strategy works as expected and results in an
approximately unbiased prediction of the spatial component.

\section{Conclusions and future work}
\label{S7}

Throughout this manuscript, we have discussed the details of SOCAIRE, the new
operational system for air quality forecasting and monitoring in the city of
Madrid. Based on an ensemble of statistical and neural models, SOCAIRE is built
under the premise that it is possible to integrate the diverse information that
correlates with air quality in order to model it. This information includes
historical values of the series itself, numerical weather and pollution
predictions, and anthropogenic features. Concretely, the proposed methodology
tackles the prediction of the four main pollutants (\no, O\textsubscript{3},
PM10, and PM2.5) for a 48-hour horizon. Thanks to its probabilistic nature, the
system is able to combine the predictions of the full probability distribution
for compound events using a Bayesian estimation of the future distribution of
the different stations over time. Thus, the system outputs are a valuable tool
for managing the \no protocol enforced by the city council of Madrid.

The tool presented in this paper is not only a theoretical proposal, but it has
been adopted as the official application to monitor, analyze and make
day-to-day decisions about air quality. The last part of this work summarizes
the structure and operation of SOCAIRE's web, as well as the main highlights of
the good results and performance of the system.

In the future, it would be interesting to apply a cost-efectiveness analysis
focused on the \no protocol activation probability. Also, we are working towards
the inclusion of a traffic forecasting system, which might improve the
performance of the models by enhancing the information that anthropogenic
features provide. Finally, SOCAIRE could be adapted to predict any kind of
combined air quality index, and not only those ones affecting the current
protocol.

\section{Acknowledgments}
\label{S8}
The authors would like to thank José Amador Fernández Viejo and the team at the
General Directorate for Sustainability of the Municipality of Madrid, especially
María de los Ángeles Cristóbal López for her continuing support and enthousiasm
towards this project.

This research has been partially funded by Empresa Municipal de Transportes
(EMT) of Madrid, Spain under the program "\textit{Aula Universitaria EMT/UNED de
  Calidad del Aire y Movilidad Sostenible}".

\printbibliography

\end{document}